\documentclass{article}
\usepackage{iclr2027_conference,times}
\usepackage[T1]{fontenc}

\usepackage{amsmath,amsfonts,bm}

\def\eqref#1{equation~\ref{#1}}

\def\1{\bm{1}}

\DeclareMathAlphabet{\mathsfit}{\encodingdefault}{\sfdefault}{m}{sl}
\SetMathAlphabet{\mathsfit}{bold}{\encodingdefault}{\sfdefault}{bx}{n}

\usepackage{hyperref}
\usepackage{url}
\usepackage{booktabs}
\usepackage{array}
\usepackage{xcolor}
\usepackage{pifont}
\usepackage{graphicx}
\usepackage{float}
\usepackage{multirow}
\usepackage{algpseudocode}

\floatstyle{ruled}
\newfloat{algorithm}{t}{loa}
\floatname{algorithm}{Algorithm}

\newcommand{\cmark}{\textcolor{green!60!black}{\ding{51}}}
\newcommand{\xmark}{\ding{55}}
\newcommand{\diaglabel}[1]{\ensuremath{\mathit{#1}}}

\title{How GUI Agents Fail: GraceBench for Execution Stability Diagnosis}

\author{
Ridong Zhang, Shuxian Liang\textsuperscript{\textdagger}, Chaoqun Wan, Zhen Huang, Xian-Sheng Hua \\[0.3em]
Institute of AI for Engineering, Tongji University \\[0.45em]
\normalfont Project page: \url{https://iaie-physicalai.github.io/GraceBench}
}

\iclrfinalcopy
\begin{document}
\raggedbottom

\maketitle
\begingroup
\renewcommand{\thefootnote}{\fnsymbol{footnote}}
\footnotetext[2]{Corresponding author.}
\endgroup

\begin{abstract}
As GUI agents are increasingly deployed for mobile task automation, their reliability depends not only on final task outcomes, but also on whether they reach these outcomes through stable interaction trajectories. Whether in daily automation scenarios or testing workflows, unnecessary actions, detours, and execution errors can add noise to collected data, introduce irreversible side effects, or undermine trust, even when the agent eventually completes the task. It is therefore important to systematically diagnose the execution stability, including whether an agent deviates from a feasible path, enters a loop, becomes stagnant, or recovers from unstable actions. Existing benchmarks, however, primarily assess final task success or a small set of intermediate milestones, leaving step-level execution behavior unclear. To bridge this gap, we introduce \textsc{GraceBench}, an online evaluation framework that represents the reachable interaction space of mobile applications as UI Transition Graphs (UTGs), providing a structural reference for step-level execution diagnosis. To support reliable alignment between dynamic UI observations and graph states, GraceBench uses video-grounded trajectory evaluation over this graph-aligned representation, assessing process stability beyond final success by measuring metrics such as deviation rate, loop burden rate, and recovery ability. GraceBench contains 169 tasks spanning 10 applications and constructs 10 UTGs with 2,909 canonical GUI states and 3,116 observed transitions. Across 3,718 trajectories, GraceBench shows that outcome scores hide execution instability: GPT-5.6 Sol achieves the highest success rate (SR) but the lowest recovery ability, 51.24\% of successful runs still contain a deviation or loop, and trajectories with the same success rate (SR) and completion rate (CR) can differ substantially in recovery behavior. Code will be made publicly available.
\end{abstract}

\section{Introduction}

Graphical user interface (GUI) agents are increasingly expected to operate real software environments on behalf of users~\citep{zhang2025largelanguagemodelbrained}, from completing routine mobile tasks~\citep{rawles2025androidworlddynamicbenchmarkingenvironment} to supporting software-testing data collection workflows.
In such settings, an agent's capability is not determined solely by whether it eventually reaches a target state; the execution process itself is equally important.
This is especially true in GUI-based automated testing, where system performance, logs, resource usage, and behavioral signals are often collected along the execution trajectory.
If an agent gets off track, repeats actions, or fails to recover, these path-dependent measurements become noisy and difficult to interpret, reproduce, and trust~\citep{dong2021flakytest}.
Execution stability is therefore a key requirement for reliable GUI-agent deployment in real interactive environments.

\begin{figure}[t]
\centering
\includegraphics[width=\linewidth]{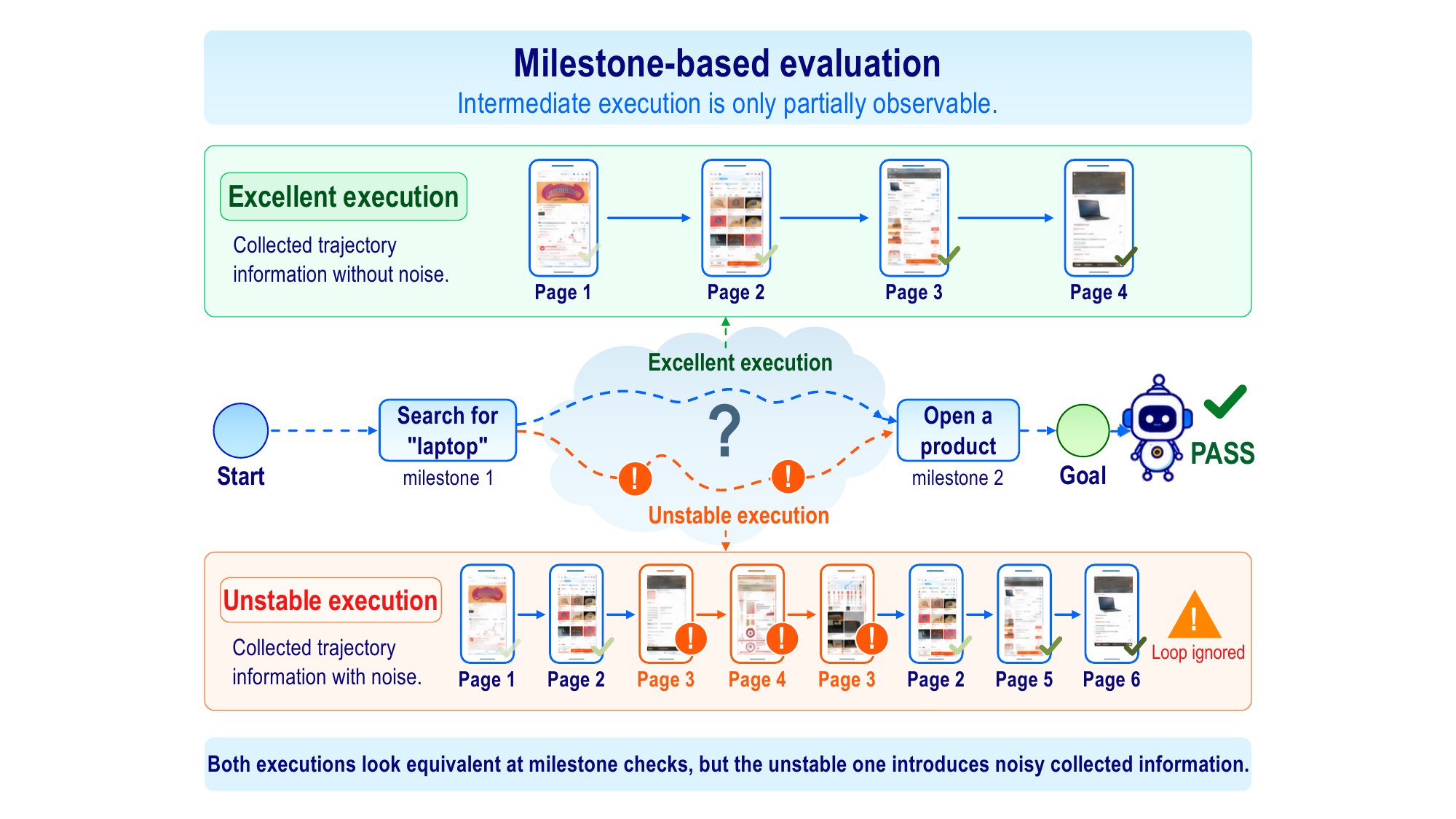}
\caption{Limitation of existing milestone-based process evaluation.
A clean execution and an unstable execution with an intermediate loop can satisfy the same semantic milestones and therefore receive the same outcome.
Observing only milestone attainment hides the loop, recovery behavior, and noise introduced along the execution trajectory.}
\label{fig:gracebench_graph_diagnosis}
\end{figure}

Existing GUI-agent benchmarks primarily evaluate task completion, usually by checking whether the final state satisfies a goal condition~\citep{rawles2025androidworlddynamicbenchmarkingenvironment}.
Although terminal success is useful, it offers little visibility into the stability of the execution process.
Some recent work introduces intermediate milestones, such as essential states or subtasks, to measure staged progress~\citep{chai2026a3androidagentarena,chen2025spabenchcomprehensivebenchmarksmartphone}, but such evaluation remains coarse-grained: it implicitly treats tasks as mostly linear and overlooks the effect of each action in a complex interactive environment.
By checking only predefined semantic milestones, milestone-based evaluation cannot assess execution stability between them.
As shown in Figure~\ref{fig:gracebench_graph_diagnosis}, existing evaluation methods cannot distinguish whether a successful agent follows a stable path, deviates and recovers, or falls into a repeated loop before completion.
Without step-level diagnostic signals, it is difficult to systematically understand execution instability in real GUI interactions.

To address this gap, we introduce \textsc{GraceBench}, a benchmark for diagnosing the execution stability of GUI agents in real mobile applications.
GraceBench uses UTGs to represent the finite reachable interaction space of an application as a graph.
In a UTG, nodes correspond to GUI states and edges correspond to executable action transitions between states~\citep{yang2015staticwindowtransitiongraphs,wen2024autodroid}.
This graph serves as a coordinate system for evaluating execution: nodes indicate where the agent is, edges indicate where the agent is moving, and the surrounding graph reveals what alternative interactions are available.
GraceBench defines each task on a task-specific UTG subgraph and maps the recorded execution trajectory back onto this graph for step-level diagnosis.
Because dynamic mobile UIs can change during and after an action, GraceBench uses synchronized video evidence to align observations with graph states and diagnose what happened between consecutive decisions.

Based on this UTG representation, GraceBench evaluates GUI agents beyond final success.
It contains 169 tasks across 10 applications and multiple domains, built on 10 UTGs with 2,909 canonical GUI states; each task's matched states and their one-hop neighborhood cover 41.09 canonical GUI states on average, providing a local structural reference for step-level diagnosis.
Each task is paired with concise and detailed instruction variants, allowing us to evaluate stability under both autonomous path planning and step-level execution guidance.
Beyond success and completion rates, GraceBench reports deviation rate, loop burden rate, and recovery ability to evaluate the execution stability of GUI agents.

We evaluate eleven models, including two GUI-specific models and nine general-purpose models, on all 169 tasks under two instruction conditions, yielding 3,718 trajectories.
Instability drives GUI-agent failures and persists even behind successful outcomes.
Across these trajectories, agents often fail through path deviation and looping, while 51.24\% of successful trajectories still contain a deviation or loop.
Even GPT-5.6 Sol, which achieves the highest success rate, has the lowest recovery ability on its observed recovery opportunities.
Outcome-controlled comparisons further show that trajectories with the same success and completion outcomes can differ substantially in recovery behavior, indicating that final success alone can overestimate GUI-agent reliability.

\begin{table}[t]
\centering
\scriptsize
\setlength{\tabcolsep}{4pt}
\caption{Comparison with representative GUI-agent benchmarks.
\textit{Venue} denotes the representative publication venue or preprint year; other columns summarize support for process evaluation, online execution, multi-path tasks, graph references, stability metrics, and video evidence.}
\label{tab:benchmark_comparison}
\resizebox{\linewidth}{!}{%
\begin{tabular}{llcccccc}
\toprule
Benchmark & Venue & Process Eval & Online Env. & Multi-Path & Graph Ref. & Stability Metrics & Video Ev. \\
\midrule
AndroidWorld~\citep{rawles2025androidworlddynamicbenchmarkingenvironment} & ICLR 2025 & \xmark & \cmark & \xmark & \xmark & \xmark & \xmark \\
SPA-Bench~\citep{chen2025spabenchcomprehensivebenchmarksmartphone} & ICLR 2025 & \cmark & \cmark & \xmark & \xmark & \xmark & \xmark \\
A3~\citep{chai2026a3androidagentarena} & Findings ACL 2026 & \cmark & \cmark & \cmark & \xmark & \xmark & \xmark \\
AndroidDaily~\citep{sui2026androiddaily} & arXiv 2026 & \cmark & \cmark & \cmark & \xmark & \xmark & \xmark \\
ProBench~\citep{yang2025probenchbenchmarkingguiagents} & AAAI 2026 & \cmark & \cmark & \xmark & \xmark & \xmark & \xmark \\
ColorBench~\citep{song2025colorbenchbenchmarkingmobileagents} & WWW 2026 & \cmark & \xmark & \cmark & \cmark & \xmark & \xmark \\
MobiFlow~\citep{feng2026mobiflowrealworldmobileagent} & arXiv 2026 & \cmark & \xmark & \cmark & \cmark & \xmark & \xmark \\
DAC-Bench~\citep{zhang2026dacbench} & ACL 2026 & \cmark & \xmark & \cmark & \xmark & \xmark & \xmark \\
GUI-RobustEval~\citep{bu2026guirecoveringpolicyinducederrors} & ICML 2026 & \cmark & \cmark & \xmark & \xmark & \cmark & \xmark \\
\midrule
\textsc{GraceBench} & This work & \cmark & \cmark & \cmark & \cmark & \cmark & \cmark \\
\bottomrule
\end{tabular}
}
\end{table}

Our contributions are summarized as follows:
\begin{itemize}
    \item We introduce \textsc{GraceBench}, a benchmark specifically designed to evaluate the execution stability of GUI agents in real mobile application environments.
    \item We propose a video-grounded online evaluation framework that maps recorded agent trajectories onto UI transition graphs, enabling step-level diagnosis of deviation, looping, and recovery ability.
    \item We systematically evaluate multiple types of GUI agents and identify shared instability patterns that are difficult to observe from final success alone, providing evidence for future diagnosis and optimization of GUI-agent capabilities.
\end{itemize}

\section{Related Work}

\subsection{Mobile GUI Agent Benchmarks}
Early GUI agent benchmarks mainly rely on offline evaluation, where predicted actions are matched against a single golden trajectory step by step~\citep{rawles2023androidwildlargescaledataset,li2024effectsdatascaleui}.
While scalable, such protocols reduce GUI control to trajectory imitation, ignoring multiple valid paths and the agent's ability to recover from deviations.
Recent online benchmarks, represented by AndroidWorld~\citep{rawles2025androidworlddynamicbenchmarkingenvironment}, allow agents to interact with executable environments and evaluate task completion from the final UI state.
However, final-success metrics remain coarse: they reveal whether a task is completed, but not how reliably it is executed.
Process-aware benchmarks further introduce intermediate supervision.
A3~\citep{chai2026a3androidagentarena} evaluates the completion ratio of essential semantic states, SPA-Bench~\citep{chen2025spabenchcomprehensivebenchmarksmartphone} verifies predefined app-level subtasks, and ProBench~\citep{yang2025probenchbenchmarkingguiagents} captures process information to check key-step completion.
These works move beyond outcome-only evaluation, but their signals are still mostly defined over milestones or semantic checkpoints.
Consequently, they provide limited visibility into failures occurring between milestones, and cannot distinguish merely successful executions from stable executions.
Our work focuses on this missing layer by evaluating GUI agents at the granularity of execution states and action transitions.

\subsection{Graph-Structured Representations for GUI Agents}
Graph structures have been widely used to model GUI interaction as transitions over discrete interface states.
Existing works use such representations for app exploration and knowledge construction~\citep{sun2025guixploreempoweringgeneralizablegui,chai2026uikobeknowledgeorientedbehaviorexploration}, task generation and pretraining~\citep{ye2025mobileagentv3fundamentalagentsgui,wu2024mobilevlmvisionlanguagemodelbetter}, and runtime planning or recovery~\citep{yu2026graphpilotguitaskautomation,guo2025agentsamastateawaremobileassistant,zhong2026actionenginereactiveprogrammaticgui}.
These methods demonstrate that graphs provide a natural abstraction for multi-step, multi-path GUI interaction.
In GUI-agent evaluation, ColorBench~\citep{song2025colorbenchbenchmarkingmobileagents} traverses UI states to construct an application-level graph, while MobiFlow~\citep{feng2026mobiflowrealworldmobileagent} builds a more abstract graph through trajectory fusion and state compression.
Both works use graph structures to make offline evaluation closer to real interaction.
However, they still overlook how dynamic changes in real environments, such as changing content and UI components, affect agent execution, whereas GraceBench aligns live execution traces with UI transition graphs to diagnose step-level deviation, stagnation, recovery, and execution stability.

\section{GraceBench}
\label{sec:gracebench}

To operationalize the graph-based diagnosis motivated above, GraceBench uses task-specific UTGs as a shared reference throughout benchmark construction and evaluation.
We first build application-level UTGs and curate 169 long-horizon tasks across 10 multi-functional mobile applications, each with concise and detailed instructions.
Agents then execute these tasks on physical devices while GraceBench records synchronized actions, observations, and video evidence.

\begin{figure}[t]
\centering
\includegraphics[width=\linewidth]{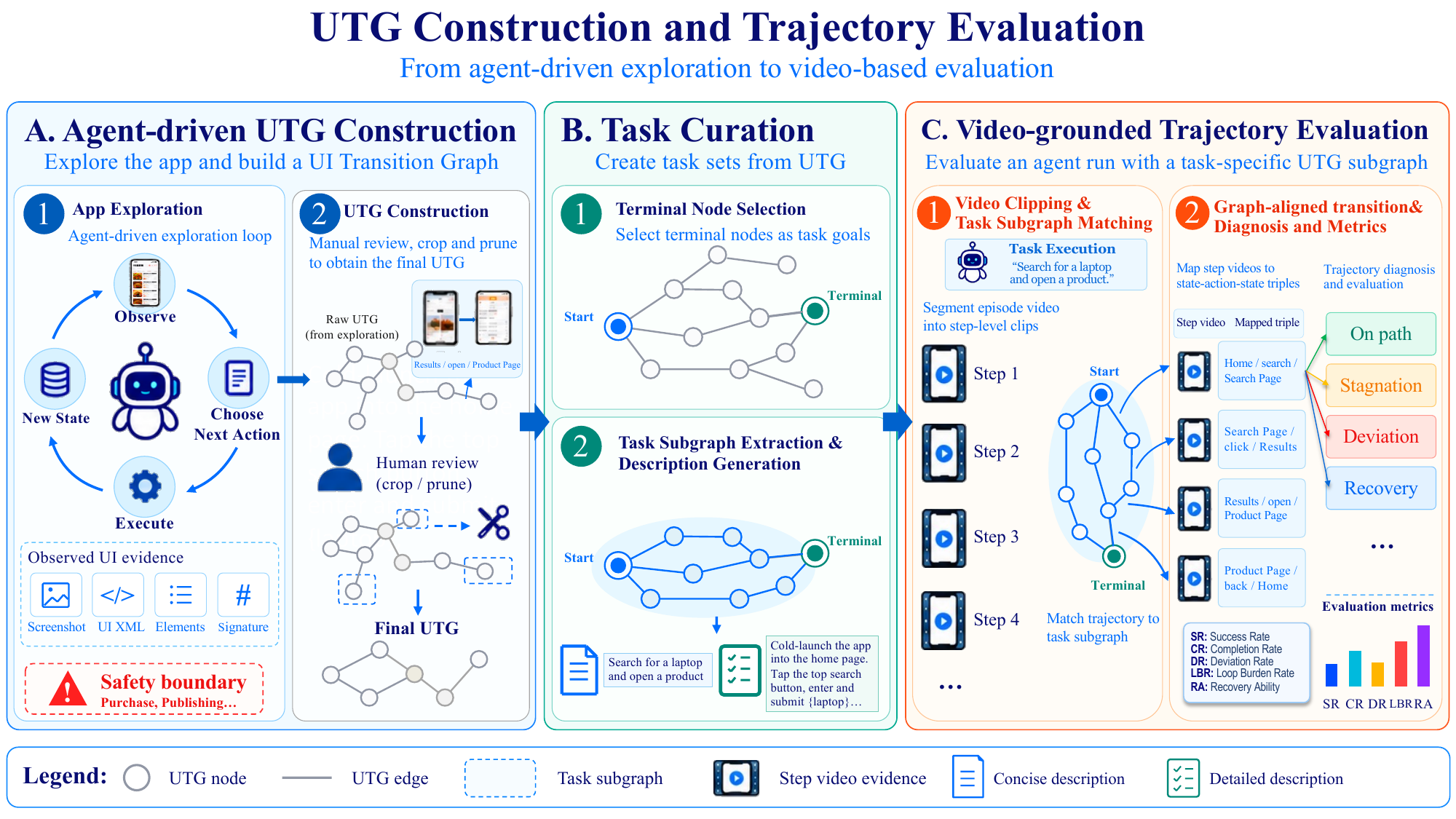}
\caption{Overview of GraceBench task curation and evaluation.
GraceBench first builds application-level UTGs through agent-driven exploration, curates task-specific subgraphs and instructions, and then evaluates online agent trajectories by mapping step-level video evidence to graph-aligned execution labels and stability metrics.}
\label{fig:gracebench_overview}
\end{figure}

Figure~\ref{fig:gracebench_overview} provides an overview of the resulting GraceBench pipeline: panel (a) illustrates agent-driven UTG construction, panel (b) summarizes task curation, and panel (c) shows video-grounded trajectory evaluation.
We describe these components in order below.

\subsection{Task Curation}
\label{sec:task_curation}

\begin{figure}[t]
\centering
\includegraphics[width=\linewidth]{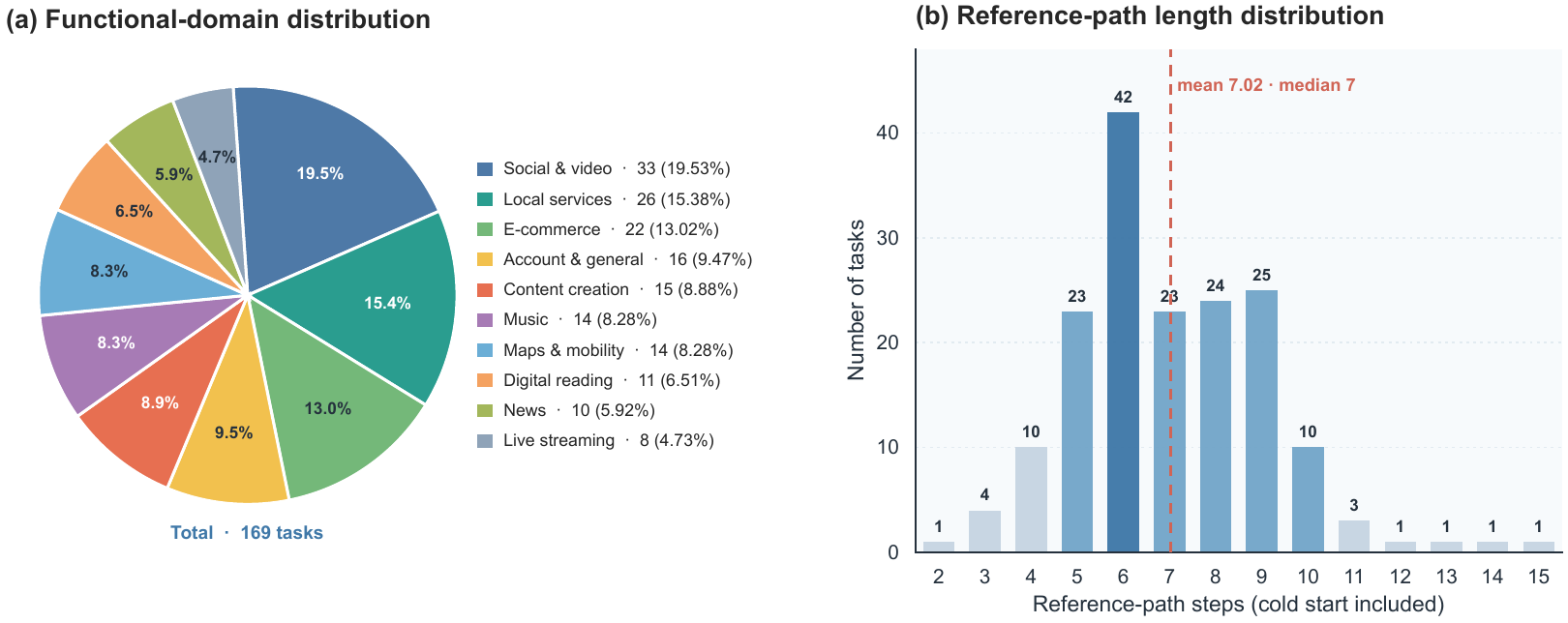}
\caption{Distribution of the 169 GraceBench tasks.
\textbf{(a)} Functional domains across 10 applications.
\textbf{(b)} Reference-path length distribution.}
\label{fig:task_domain_distribution}
\end{figure}

\paragraph{UTG construction.}
For each application, we construct an application-level UTG as the structural basis for task curation and trajectory evaluation.
We denote an application-level UTG as \(G=(V,E)\), where \(V\) is the set of task-relevant GUI states and \(E\) is the set of directed action transitions between states.
Each node \(v\in V\) corresponds to a visually verifiable GUI state, and each edge \(e=(v_i,a,v_j)\) indicates that executing action \(a\) at state \(v_i\) leads to state \(v_j\).

We use an agent-assisted exploration loop to collect UI transitions.
Before exploration, we set the application scope and treat high-risk operations such as payment, order submission, and authorization changes as boundary states.
At each step, the agent observes the current screen through ADB, including both the screenshot and UI hierarchy, identifies interactable elements, selects a candidate action, and executes the action through ADB~\citep{androiddevelopers2026adb}.
Each observed interface transition is recorded as a state-action-state edge in the UTG.
Detailed implementation of the automatic exploration loop is provided in Appendix~\ref{app:utg_exploration}.

After exploration, we clean and normalize the collected graph by deduplicating equivalent UI states, filtering unstable textual attributes, and saving a representative screenshot for each retained state.
Across 10 UTGs, GraceBench normalizes 3,265 collected state observations into 2,909 canonical GUI states, together with their corresponding observed transitions.
Appendix~\ref{app:utg_structural_coverage} reports the per-application inventory and counting rules.
All UTGs used in GraceBench are manually reviewed before task extraction to ensure state validity, transition executability, and coverage of major functional pathways.

\paragraph{Task-subgraph extraction.}
From the manually reviewed UTGs, we choose visually verifiable terminal states with stable completion semantics as task goals, excluding shallow entry or intermediate-panel states.
For each task \(\tau\), we define a natural initial state \(v_{\tau}^{\mathrm{start}}\) and target state \(v_{\tau}^{\mathrm{goal}}\), then extract a task-specific subgraph \(G_{\tau}=(V_{\tau},E_{\tau})\subseteq G\) containing task-relevant states and executable transitions that make legal progress toward the goal.
This subgraph includes alternative valid routes but excludes task-irrelevant detours, providing a multi-path structural reference for trajectory diagnosis and process-level metrics.

\paragraph{Instruction generation.}
For each task, GraceBench derives two instruction variants from the task subgraph: a \emph{concise instruction} that states the user goal with necessary constraints, and a \emph{detailed instruction} that follows one manually verified feasible path with step-level operations.
This paired design varies instruction granularity while keeping the task goal fixed, separating planning ambiguity from low-level execution stability.
We further parameterize query terms, categories, target entities, and selectors to make tasks extensible and reduce overfitting to fixed surface forms.

\subsection{Online Execution Environment}
\label{sec:online_execution}

Following ProBench~\citep{yang2025probenchbenchmarkingguiagents}, GraceBench evaluates agents on physical Android devices in an online loop.
At each step, the agent receives the current screenshot, task instruction, and interaction history, and its textual action is parsed into ADB commands for execution until completion or budget exhaustion.
Each episode is recorded as
\begin{equation}
T=(o_1,a_1,o_2,a_2,\ldots,o_T,a_T,o_{T+1}),
\end{equation}
where \((o_t,a_t,o_{t+1})\) denotes the pre-action screen, executed action, and stabilized post-action screen used for step-level evaluation.

GraceBench uses the action space \texttt{CLICK}, \texttt{SWIPE}, \texttt{TYPE}, \texttt{ENTER}, \texttt{BACK}, \texttt{COMPLETE}, and \texttt{WAIT}; \texttt{HOME} is excluded because tasks stay within one application.

Video capture is integrated into the same execution loop.
GraceBench starts a continuous screen recording before the first agent decision and stops it after the episode terminates.
The recording is implemented with \texttt{scrcpy}~\citep{genymobile2026scrcpy}, while the runner logs the action type, action timestamp, pre-action screenshot, and post-action screenshot for each step.
This ensures that the video evidence is synchronized with the agent's actual decision and device-action timeline.

Prior screencast-based GUI analysis uses video to preserve the temporal process from user operation to visual feedback~\citep{liu2025screencast}.
GraceBench instead uses video as temporal evidence for dynamic GUI effects such as loading, pop-ups, animations, and content changes, which motivates the graph-aligned diagnosis below.

\subsection{Video-Grounded Trajectory Evaluation}
\label{sec:trajectory_evaluation}

\begin{figure}[t]
\centering
\includegraphics[width=\linewidth]{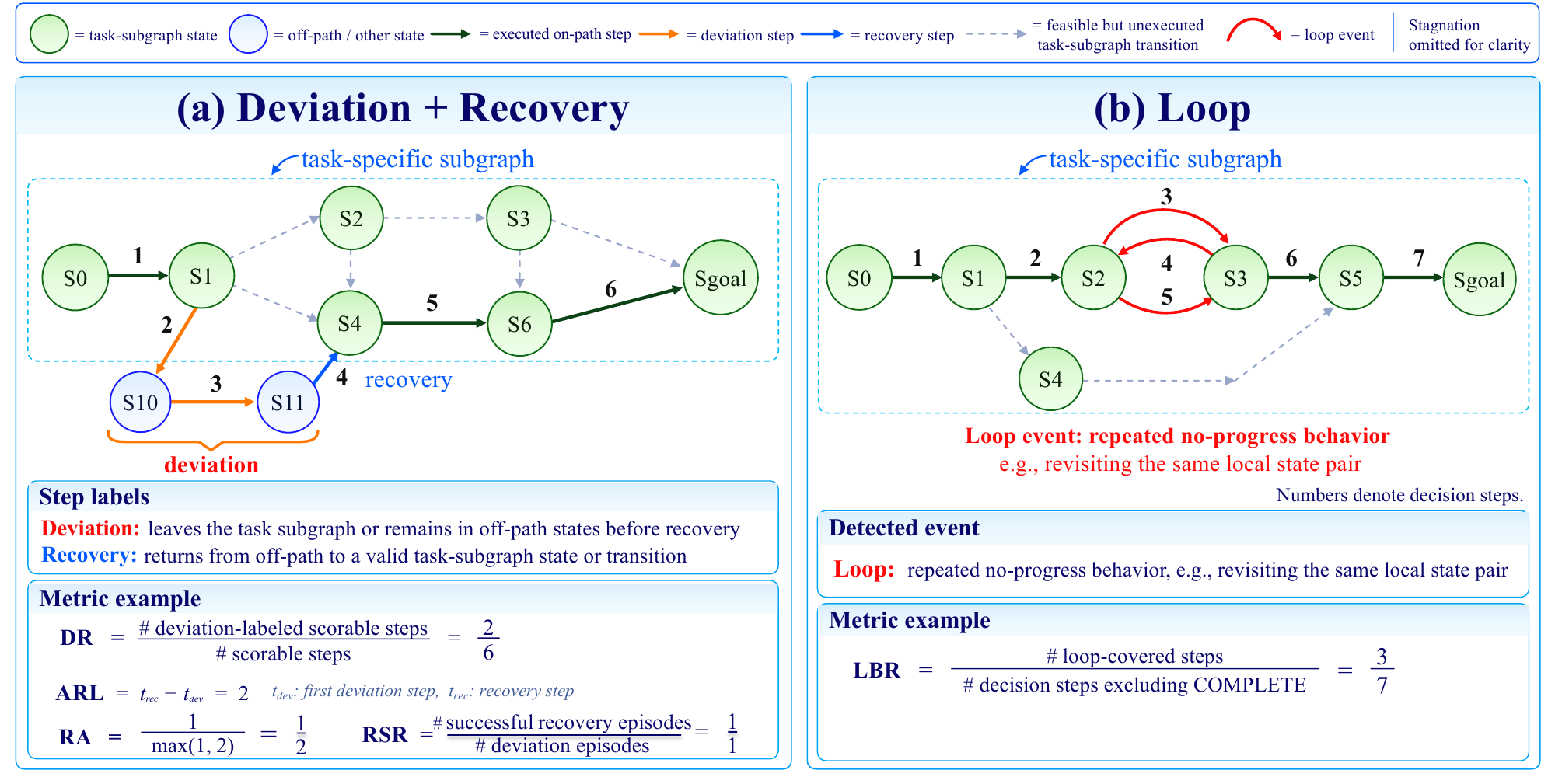}
\caption{Trajectory-diagnosis examples on task-specific UTGs.
\textbf{(a)} Steps 2--3 leave the task subgraph and step 4 recovers, providing the labels used to compute DR, ARL, RA, and RSR.
\textbf{(b)} Repeated transitions between the same local state pair form a loop event and determine LBR.
Dashed edges denote feasible but unexecuted task-subgraph transitions; stagnation is omitted for clarity.}
\label{fig:trajectory_diagnosis_showcase}
\end{figure}

GraceBench evaluates each execution trace by aligning observed agent behavior with the task subgraph.
The evaluation pipeline contains four stages.
First, the episode-level video is segmented into step-level clips based on action timestamps.
These clips provide visual evidence for what happened before, during, and after each action.
Second, GraceBench retrieves the task-specific subgraph from the full application UTG.
The subgraph contains the task initial state, target state, required transitions, and alternative valid paths, serving as the reference structure for judging whether an observed action advances the task.
Third, a video-based evaluator analyzes each local transition \((o_t,a_t,o_{t+1})\) using the corresponding step-level video clip and executed action.
The evaluator maps the pre-action and post-action observations to graph states.
For each observation, the matched graph state is defined as
\begin{equation}
\hat{v}_t =
\begin{cases}
v_i, & \text{if } o_t \text{ matches a state } v_i \in V,\\
v_{\mathrm{out}}, & \text{otherwise},
\end{cases}
\end{equation}
where \(v_{\mathrm{out}}\) denotes a state outside the application-level UTG.
This mapping converts each local transition into a graph-aligned transition:
\begin{equation}
(o_t,a_t,o_{t+1}) \mapsto (\hat{v}_t,a_t,\hat{v}_{t+1}).
\end{equation}
Based on the aligned transition, the evaluator assigns one of four step-level labels: \diaglabel{on\ path}, \diaglabel{stagnation}, \diaglabel{deviation}, or \diaglabel{recovery}.
A step is labeled \diaglabel{on\ path} if it follows a valid transition in \(G_{\tau}\), \diaglabel{stagnation} if it causes no meaningful state change, \diaglabel{deviation} if it leaves the task subgraph or enters \(v_{\mathrm{out}}\), and \diaglabel{recovery} if it returns from an off-path state to a valid state or transition in \(G_{\tau}\).
The resulting labeled sequence is the graph-aligned diagnostic trajectory:
\begin{equation}
T'=[(\hat{v}_1,a_1,\hat{v}_2,\ell_1),\ldots,(\hat{v}_T,a_T,\hat{v}_{T+1},\ell_T)].
\end{equation}
GraceBench uses \(T'\) to identify trajectory-level behaviors such as missing required transitions, persistent deviation, repeated no-progress actions, loops, successful recovery after deviation, and how unmatched \(v_{\mathrm{out}}\) states should be behaviorally labeled.
The step-level diagnostic labels are first generated by expert agents and then validated on a preregistered, stratified 10\% sample against blinded human expert labels (Appendix~\ref{app:empirical_validation}).
The audit estimates agreement between generated labels and human expert judgments on sampled transitions; it does not replace every generated label in the full dataset.
Figure~\ref{fig:trajectory_diagnosis_showcase} illustrates how the graph alignment localizes deviation, recovery, and loop events and connects them to the corresponding process metrics.

\subsection{Evaluation Metrics}
\label{sec:evaluation_metrics}

\paragraph{Task evaluation metrics.}
At the task level, we report task success and graph-aligned execution quality.
(1) \emph{SR} measures the fraction of trajectories marked successful in the per-trajectory evaluation summaries; these labels are preserved in all reported analyses.
(2) \emph{CR} measures how much of the task-required subgraph has been completed, providing a finer-grained view of progress when the final task fails.
(3) \emph{Deviation rate (DR)} measures how often the agent moves outside the task subgraph, capturing path instability.
(4) \emph{Loop burden rate (LBR)} measures how much execution is consumed by repeated no-progress behavior once looping occurs.
(5) \emph{Recovery success rate (RSR)} measures whether the agent can return to the task subgraph after a deviation.
(6) \emph{Average recovery latency (ARL)} measures how many steps successful recovery takes.
(7) \emph{Recovery ability (RA)} combines recovery success and recovery speed into a single robustness score.
Detailed calculation rules are provided in Appendix~\ref{app:metric_details}.

\section{Experiments}
\label{sec:experiments}
\textbf{Our experiments} address three questions: \emph{How do GUI agents fail?} \emph{Do task outcomes adequately characterize execution stability?} Finally, \emph{Do stability metrics capture reproducible agent differences across tasks?}
After the setup, each subsection answers one question using outcome-stratified statistics, outcome-controlled comparisons, or cross-task replication.
Table~\ref{tab:main_results} summarizes the main experimental results across agents, instruction conditions, and stability metrics.

\subsection{Experimental Setup}
\paragraph{Baseline agents.}
We evaluate representative screenshot-only GUI agents, including GUI-specific models and general-purpose models from the DeepSeek, Gemini, GPT, Qwen, and GLM families, using each model's official prompting and tool-calling setup when available.
Each model is tested under detailed subgraph-derived instructions and concise user-level goals, with temperature set to 0 and a maximum budget of 15 actions.
Table~\ref{tab:main_results} contains 3,718 trajectories from eleven completed models, two instruction conditions, and 169 tasks.
All trajectory-level analyses use this full eleven-model dataset; paired instruction analyses contain 1,859 model--task pairs.
The two GUI-specific models use 8-bit \texttt{MLX} checkpoints~\citep{apple2026mlx} and were collected on a separate host; their evaluated records are included in every applicable analysis.
In addition to SR and CR, we report DR, LBR, RSR, ARL, and RA to evaluate GUI-agent execution stability; detailed definitions and aggregation rules are provided in Appendix~\ref{app:metric_details}.
\begin{table}[t]
\centering\scriptsize
\setlength{\tabcolsep}{3pt}
\renewcommand{\arraystretch}{1.08}
\caption{Results for all 3,718 trajectories: eleven models, 169 tasks, and two instruction conditions. All rows use the supplied per-trajectory evaluation summaries, preserving the reported SR labels and process scores. Data provenance is detailed in Appendix~\ref{app:diagnostic_ablations}. Rates are decimals; ARL is measured in steps. Within each instruction block, best values are bolded and second-best values are underlined.}
\label{tab:main_results}
\resizebox{\linewidth}{!}{%
\begin{tabular}{@{}llccccccc@{}}
\toprule
\multicolumn{9}{@{}l}{\textbf{Detailed instruction}} \\
\midrule
Type & Model & SR$\uparrow$ & CR$\uparrow$ & DR$\downarrow$ & LBR$\downarrow$ & RA$\uparrow$ & RSR$\uparrow$ & ARL$\downarrow$ \\
\midrule
\multirow{2}{*}{\textit{GUI-specific}} & GUI-Owl-1.5-8B-Instruct & 0.1657 & 0.4403 & 0.3306 & 0.5559 & \underline{0.6603} & \textbf{0.8212} & 1.6877 \\
 & UI-TARS-1.5-7B & 0.1006 & 0.2956 & 0.5149 & 0.5133 & 0.6017 & 0.7888 & 1.9501 \\
\midrule
\multirow{5}{*}{\textit{General open-source}} & DeepSeek-V4.1-Flash & 0.5148 & 0.6099 & 0.1348 & 0.5934 & 0.5602 & 0.6987 & 1.8716 \\
 & GLM-5.3-Flash & \underline{0.5976} & \underline{0.7304} & \textbf{0.0743} & \underline{0.4470} & 0.5605 & 0.6857 & 1.4583 \\
 & Qwen3.8-Flash & 0.3254 & 0.4474 & 0.1591 & 0.6606 & 0.5834 & 0.7819 & 1.7755 \\
 & Qwen3.8-Max & 0.2722 & 0.4389 & 0.2111 & 0.6945 & 0.6017 & 0.7979 & 1.8400 \\
 & Qwen3.8-27B & 0.2367 & 0.4044 & 0.2286 & 0.7412 & 0.5703 & 0.6976 & 1.5780 \\
\midrule
\multirow{4}{*}{\textit{Proprietary}} & GPT-5.6 Sol & \textbf{0.6391} & \textbf{0.7388} & \underline{0.1223} & \textbf{0.3870} & 0.4454 & 0.5645 & 1.7143 \\
 & GPT-5.5 & 0.5207 & 0.6151 & 0.1855 & 0.4502 & 0.5061 & 0.5928 & \underline{1.4141} \\
 & Gemini 3.5 Flash & 0.3373 & 0.5428 & 0.1930 & 0.5278 & \textbf{0.6988} & \underline{0.8106} & \textbf{1.3689} \\
 & Gemini 3.1 Flash Lite & 0.0710 & 0.1799 & 0.3961 & 0.7277 & 0.5238 & 0.7269 & 2.2597 \\
\midrule
\multicolumn{9}{@{}l}{\textbf{Concise instruction}} \\
\midrule
Type & Model & SR$\uparrow$ & CR$\uparrow$ & DR$\downarrow$ & LBR$\downarrow$ & RA$\uparrow$ & RSR$\uparrow$ & ARL$\downarrow$ \\
\midrule
\multirow{2}{*}{\textit{GUI-specific}} & GUI-Owl-1.5-8B-Instruct & 0.2367 & 0.4393 & 0.3807 & 0.5041 & 0.6440 & \underline{0.8296} & 1.9366 \\
 & UI-TARS-1.5-7B & 0.1538 & 0.3643 & 0.5088 & \underline{0.4572} & 0.6013 & 0.8000 & 1.9740 \\
\midrule
\multirow{5}{*}{\textit{General open-source}} & DeepSeek-V4.1-Flash & 0.4142 & 0.5163 & 0.2128 & 0.6107 & 0.6020 & 0.8201 & 1.9388 \\
 & GLM-5.3-Flash & \underline{0.6272} & \textbf{0.6676} & \textbf{0.1303} & \textbf{0.4289} & 0.5560 & 0.7070 & 1.7027 \\
 & Qwen3.8-Flash & 0.3787 & 0.5436 & 0.2142 & 0.5508 & \textbf{0.7193} & \textbf{0.8581} & \underline{1.5269} \\
 & Qwen3.8-Max & 0.3077 & 0.4035 & 0.2154 & 0.7103 & \underline{0.6867} & 0.8087 & \textbf{1.4523} \\
 & Qwen3.8-27B & 0.1893 & 0.3360 & 0.2907 & 0.7532 & 0.5410 & 0.6835 & 1.6158 \\
\midrule
\multirow{4}{*}{\textit{Proprietary}} & GPT-5.6 Sol & \textbf{0.6391} & \underline{0.6261} & \underline{0.1856} & 0.4746 & 0.4525 & 0.5806 & 1.8111 \\
 & GPT-5.5 & 0.4970 & 0.5224 & 0.2753 & 0.5154 & 0.4881 & 0.5971 & 1.8374 \\
 & Gemini 3.5 Flash & 0.3077 & 0.5030 & 0.2462 & 0.5201 & 0.6654 & 0.8035 & 1.5971 \\
 & Gemini 3.1 Flash Lite & 0.1243 & 0.1902 & 0.4650 & 0.7190 & 0.4565 & 0.6150 & 2.2734 \\
\bottomrule
\end{tabular}}
\end{table}
\subsection{GUI Agents Fail Through Unstable Execution}
\label{sec:failure_patterns}
\paragraph{Failure combines incomplete progress with unstable execution.}
Table~\ref{tab:main_results} shows that GPT-5.6 Sol and GLM-5.3-Flash are the strongest overall agents: GPT-5.6 Sol attains the highest SR in both instruction settings (0.6391), while GLM-5.3-Flash achieves the lowest DR in both settings (0.0743 detailed, 0.1303 concise) and the highest concise CR (0.6676).
Yet strong outcomes do not imply uniformly stable execution.
Even GPT-5.6 Sol has the lowest reported RA on its observed recovery opportunities under both instruction settings, showing that high final success can coexist with weak recovery after mistakes.
Other agents expose different failure profiles: UI-TARS-1.5-7B is dominated by deviation (DR 0.5149 detailed, 0.5088 concise), whereas Qwen3.8-27B shows especially high loop burden (LBR 0.7412 detailed, 0.7532 concise).

\paragraph{Failed executions are less stable throughout the trajectory.}
Across all 3,718 trajectories, failures have higher DR (0.3067 vs. 0.0889 detailed; 0.3618 vs. 0.1412 concise), higher loop incidence (0.6525 vs. 0.2066; 0.6520 vs. 0.2748), higher LBR after looping begins (0.6390 vs. 0.4257; 0.6142 vs. 0.4840), and lower RA (0.5668 vs. 0.7285; 0.5918 vs. 0.6394).
Thus task failure is closely associated with sustained off-path and repeated no-progress behavior; the full outcome-stratified comparison is reported in Appendix Table~\ref{tab:outcome_process_comparison}.

\begin{figure}[!t]
\centering
\includegraphics[width=\linewidth]{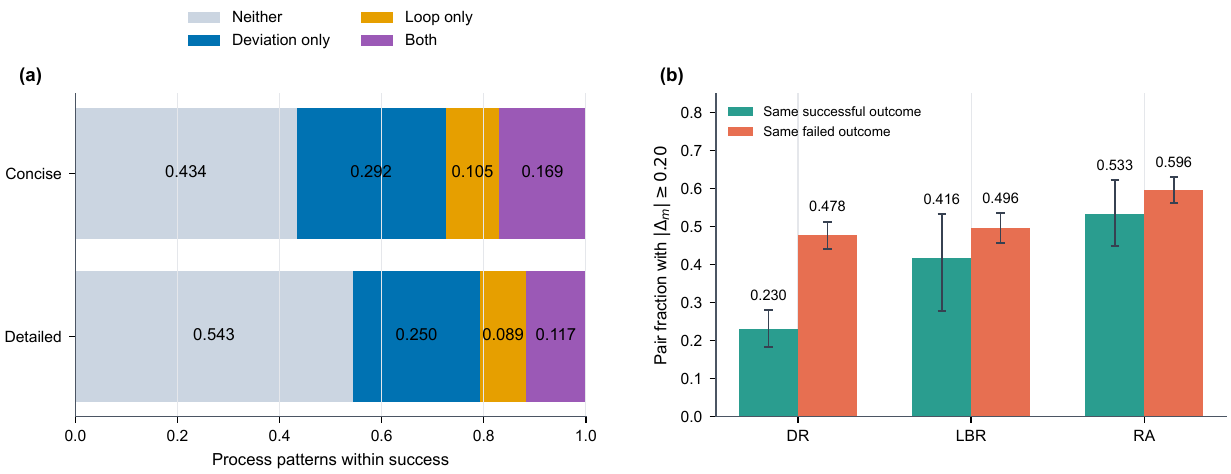}
\caption{\textbf{Task outcomes hide execution instability.} (a) Mutually exclusive process-pattern proportions among successful trajectories. (b) Exact-CR matched pairs with process gap at least 0.2000; bars distinguish successful and failed pairs, and whiskers show task-cluster 95\% intervals.}
\label{fig:complete_success}
\end{figure}

\subsection{Stability Metrics Reveal Differences Beyond Task Outcomes}
\label{sec:outcome_blindness}

\paragraph{Successful outcomes still contain unstable execution.}
Although failed trajectories are less stable on average, instability is not confined to unsuccessful runs.
Among the 1,294 successful trajectories, 51.24\% (663/1,294) contain at least one deviation or loop.
Under detailed instructions, 36.78\% of successful trajectories contain deviations and 20.66\% contain loops; under concise instructions, the corresponding rates increase to 46.11\% and 27.48\%.
Figure~\ref{fig:complete_success}(a) separates deviation-only, loop-only, and combined patterns within these successful trajectories.
Moreover, 126 successful trajectories have zero DR but still contain a loop, showing that even an on-path execution can waste actions through repeated no-progress behavior.
These results show that successful outcomes can still mask deviations and loops, so final success alone does not establish stable task execution; the outcome relevance of timely recovery after deviation is examined separately in Appendix~\ref{app:recovery_outcome_relevance}.

\paragraph{Equivalent task performance masks substantial stability differences.}
Success is only the coarsest form of outcome equivalence.
We therefore compare models executing the same task under the same instruction condition, with identical SR and equal stored CR.
Among eligible successful pairs, DR, LBR, and RA differ by at least 0.2000 in fractions 0.2305, 0.4158, and 0.5331, respectively.
Figure~\ref{fig:complete_success}(b) shows that these large process gaps also appear in failed pairs, with RA differing by at least 0.2000 in 0.5956 of exact-CR matched failed pairs.
Matching sensitivity and partially overlapping metric correlations provide additional support for this distinction (Appendix Figure~\ref{fig:diagnostic_matched_pairs}).
Thus even matched task outcomes underdetermine both the degree and the form of execution instability.

\subsection{Stability Metrics Capture Consistent Agent Differences}
\label{sec:model_consistency}
\paragraph{Model behavior profiles persist across tasks.}
We test whether process metrics capture reproducible model behavior by repeatedly splitting the 169 tasks into stratified halves, ranking all eleven models by each metric on each half, and measuring rank agreement with Spearman $\rho$.
The strongest consistency appears for DR and LBR: detailed-instruction rankings reach \(\rho=0.9516\) and \(0.9284\), while concise-instruction rankings remain high at \(\rho=0.9228\) and \(0.8381\).
RSR is also stable across splits (\(\rho=0.8100\) detailed; \(0.8269\) concise), whereas ARL is weaker (\(\rho\approx0.58\)) and RA is condition-dependent.
All metric--condition correlations remain significant after Bonferroni correction (\(p<0.05\)); the full table and permutation details are reported in Appendix Table~\ref{tab:model_consistency}.

\section{Conclusion}

We introduced \textsc{GraceBench}, which aligns video-recorded trajectories with task-specific UI Transition Graphs to diagnose deviation, looping, and recovery during real mobile execution.
Across eleven models and 3,718 trajectories, GraceBench reveals execution weaknesses hidden by SR and CR: lower-performing agents deviate and loop more often, while even the highest-SR model exhibits limited recovery on observed recovery opportunities.
Successful trajectories, as well as trajectories with identical SR and CR, can still differ substantially in DR, LBR, and RA, showing that reliable GUI automation requires more than task completion alone.
By providing graph-aligned evidence and process-level diagnostics, GraceBench supports future work on stronger state tracking, loop detection, and recovery policies for GUI agents.

\subsection*{AI use statement}

In this work, we used generative AI tools to assist benchmark construction, including drafting candidate task instructions, expanding parameterized task variants, and producing auxiliary transition annotations for GUI trajectory diagnosis.
All AI-assisted outputs were treated as candidate artifacts rather than ground truth.
The authors manually reviewed the retained UTGs, task definitions, task instructions, and screenshots to ensure state validity, transition executability, task feasibility, and consistency with the benchmark protocol.
For step-level diagnostic labels, expert-agent outputs were not assumed to be fully human-labeled; instead, we conducted a preregistered, stratified audit in which sampled transition labels were validated against blinded human expert judgments, as described in Appendix~\ref{app:empirical_validation}.
The reported quantitative results are computed from the documented evaluation records and metric definitions, and generative AI tools were not used to make final scientific claims without author verification.
The authors take full responsibility for the final content of this work, including text, claims, data, and artifacts produced with the aid of generative AI.

\subsection*{Ethics statement}

GraceBench is designed to diagnose GUI-agent execution stability in controlled mobile environments.
The benchmark does not involve human-subject experiments.
Because GUI agents may interact with real applications, we constrain tasks to avoid irreversible or high-risk operations such as payment, order submission, purchases, account changes, or authorization changes.
During UTG construction and evaluation, such operations are treated as boundary states and are not executed.
Screenshots, videos, and traces are collected from controlled devices and benchmark accounts, and released artifacts will be reviewed to remove or anonymize sensitive information where necessary.
The benchmark is intended for reliability diagnosis and safety-oriented analysis rather than for enabling harmful automation.

\subsection*{Reproducibility statement}

To support reproducibility, we will release the benchmark task definitions, instruction variants, UTG metadata, evaluation code, metric definitions, and trace/evaluation schemas, with detailed metric calculation rules provided in the appendix. For each evaluated run, GraceBench stores action traces, pre-action and post-action screenshots, action metadata, timing information, and synchronized video evidence, and we document the evaluation environment so that trajectory judgments can be audited despite changes in mobile applications, app versions, accounts, devices, and online content. The current analysis package contains per-trajectory evaluation records for all eleven models; the two GUI-specific models were collected on a separate host, and their video and screenshot files are not included in this package.

\bibliography{references}
\bibliographystyle{iclr2027_conference}

\clearpage
\appendix
\clearpage
\appendix

\section{Task-to-Domain Mapping}
\label{app:domain_mapping}

\begin{table}[H]
\centering
\small
\setlength{\tabcolsep}{4pt}
\caption{Task-to-domain mapping used in Figure~\ref{fig:task_domain_distribution}.
Tasks are grouped by their primary user goal.
Parenthesized names denote scenario-specific subgraphs within an application; the six Douyin subgraphs remain components of one Douyin application-level UTG.}
\label{tab:task_domain_mapping}
\begin{tabular}{>{\raggedright\arraybackslash}p{0.24\linewidth} >{\raggedright\arraybackslash}p{0.49\linewidth} >{\raggedleft\arraybackslash}p{0.08\linewidth} >{\raggedleft\arraybackslash}p{0.10\linewidth}}
\toprule
Task domain & Application/scenario assignment & Tasks & Share \\
\midrule
Social \& Video Entertainment & TikTok; Xiaohongshu; Hongguo Short Drama & 33 & 19.53\% \\
Local Services & Douyin (Group Buying); Dianping & 26 & 15.38\% \\
E-commerce & Douyin (Mall); Douyin Mall & 22 & 13.02\% \\
Account \& General Services & Douyin (Home Sidebar; Profile) & 16 & 9.47\% \\
Content Creation & Douyin (Content Creation) & 15 & 8.88\% \\
Music & Qishui Music & 14 & 8.28\% \\
Maps \& Mobility & Amap (Gaode Map) & 14 & 8.28\% \\
Digital Reading & Fanqie Novel & 11 & 6.51\% \\
News \& Information & Toutiao & 10 & 5.92\% \\
Live Streaming & Douyin (Live) & 8 & 4.73\% \\
\midrule
\textbf{Total} & \textbf{10 applications / 10 application-level UTGs} & \textbf{169} & \textbf{100.00\%} \\
\bottomrule
\end{tabular}
\end{table}

\section{Metric Calculation Details}
\label{app:metric_details}

This appendix gives the detailed calculation rules for the metrics used in GraceBench.
Let \(G_\tau=(V_\tau,E_\tau)\) be the task subgraph for task \(\tau\), and let \(E_{\mathrm{req}}\subseteq E_\tau\) be the set of required task transitions.
After evaluating an execution trace, GraceBench identifies the set of missing required edges \(M_\tau\subseteq E_{\mathrm{req}}\), and the number of completed required edges is
\begin{equation}
C_{\mathrm{req}}^\tau = |E_{\mathrm{req}}| - |M_\tau|.
\end{equation}
Let \(\mathcal{I}_\tau\) denote the scorable action steps in the aligned trajectory, excluding \texttt{WAIT} and \texttt{COMPLETE}.

\paragraph{Success Rate.}
Success rate is the proportion of runs marked successful in the frozen per-trajectory evaluation summaries.
For each task,
\begin{equation}
SR_\tau = \mathbf{1}[\text{the evaluation summary reports task success}],
\end{equation}
The task-success labels are taken directly from the supplied summaries and preserved throughout aggregation and post-hoc analysis.
The benchmark-level success rate is
\begin{equation}
SR = \frac{1}{N}\sum_{\tau=1}^{N} SR_\tau .
\end{equation}

\paragraph{Completion Rate.}
Completion rate measures the fraction of task-required transitions completed by the agent:
\begin{equation}
CR_\tau = \frac{C_{\mathrm{req}}^\tau}{|E_{\mathrm{req}}|}.
\end{equation}
We report macro-averaged CR over all evaluated tasks.

\paragraph{Deviation Rate.}
Deviation rate measures the fraction of scorable steps labeled as \diaglabel{deviation}:
\begin{equation}
DR_\tau = \frac{|\{t\in \mathcal{I}_\tau : \ell_t=\diaglabel{deviation}\}|}{|\mathcal{I}_\tau|}.
\end{equation}
We report macro-averaged DR over all evaluated tasks.

\paragraph{Loop Burden Rate.}
Let \(L_\tau\) be the set of steps covered by detected loop events in task \(\tau\).
Let \(\mathcal{D}_\tau\) denote all decision steps except \texttt{COMPLETE}; unlike \(\mathcal{I}_\tau\), it includes \texttt{WAIT}, since repeated waiting can consume the interaction budget.
For a trajectory with at least one loop, the loop step rate is
\begin{equation}
LBR_\tau = \frac{|L_\tau|}{|\mathcal{D}_\tau|}.
\end{equation}
In the main results, loop burden is aggregated only over trajectories where a loop occurs:
\begin{equation}
LBR = \frac{\sum_{\tau: |L_\tau|>0} |L_\tau|}{\sum_{\tau: |L_\tau|>0} |\mathcal{D}_\tau|}.
\end{equation}

\paragraph{Recovery Success Rate.}
A recovery opportunity begins when the agent enters an off-path state.
For task \(\tau\), let \(N_{\mathrm{dev}}^\tau\) be the number of deviation episodes and \(N_{\mathrm{succ}}^\tau\) be the number of episodes that successfully return to the task subgraph.
When \(N_{\mathrm{dev}}^\tau>0\),
\begin{equation}
RSR_\tau = \frac{N_{\mathrm{succ}}^\tau}{N_{\mathrm{dev}}^\tau}.
\end{equation}
Tasks without recovery opportunities are excluded from aggregate recovery analysis.
The reported RSR pools successful episodes and opportunities across applicable tasks, rather than averaging their individual ratios.

\paragraph{Average Recovery Latency.}
For each successful recovery episode \(i\), recovery latency is the number of steps from the first deviation step to the step where the trajectory returns to the task subgraph:
\begin{equation}
\mathrm{latency}_i = t_i^{\mathrm{rec}} - t_i^{\mathrm{dev}}.
\end{equation}
ARL is averaged over successful recovery episodes:
\begin{equation}
ARL = \frac{1}{N_{\mathrm{succ}}}\sum_{i=1}^{N_{\mathrm{succ}}} \mathrm{latency}_i .
\end{equation}

\paragraph{Recovery Ability.}
Recovery ability combines recovery success and recovery speed.
For each recovery opportunity \(i\), the contribution is
\begin{equation}
c_i =
\begin{cases}
\frac{1}{\max(1,\mathrm{latency}_i)}, & \text{if recovered},\\
0, & \text{otherwise}.
\end{cases}
\end{equation}
For a task with \(N_{\mathrm{dev}}^\tau>0\),
\begin{equation}
RA_\tau = \frac{1}{N_{\mathrm{dev}}^\tau}\sum_{i=1}^{N_{\mathrm{dev}}^\tau} c_i .
\end{equation}
As with RSR, tasks without recovery opportunities are treated as not applicable rather than assigned a perfect recovery score.
The reported RA pools the contributions \(c_i\) across all episodes and divides by the total recovery-opportunity count.

\section{Post-hoc Diagnostic Ablations}
\label{app:diagnostic_ablations}
\paragraph{Dataset scope and reproducibility.}
Every trajectory-level analysis in the main text and this appendix starts from the same 3,718 records: eleven models, 169 tasks, and two instruction conditions.
Outcome and metric-availability restrictions define the explicitly reported supports within this full dataset.
The completed reruns of 78 error-ended and 10 invalid-startup cases are included under the same reconciliation rules used for the main analysis.
The existing nine-model summaries contribute 3,042 records.
The imported UI-TARS and GUI-Owl bundle contributes the remaining 676 records.
All task identities, instructions, process metrics, count denominators, and reported SR labels are retained from the supplied per-trajectory summaries.
No labels are replaced using separate terminal reviews.
The imported package does not contain video or screenshot bytes, so this analysis reuses existing evaluations and does not perform new visual adjudication.
The remainder of this appendix provides the supporting analyses referenced in Section~\ref{sec:experiments}, including outcome-stratified instability, instruction effects, outcome-controlled comparisons, cross-task model consistency, recovery relevance, and predictive-cost ablations.

\paragraph{Outcome-stratified process instability.}
Table~\ref{tab:outcome_process_comparison} provides the full comparison summarized in Section~\ref{sec:failure_patterns}.
\begin{table}[t]
\centering\scriptsize
\setlength{\tabcolsep}{4pt}
\caption{Process instability by final task outcome across all 3,718 trajectories. Gaps are unsuccessful minus successful. DR is averaged over all trajectories; loop incidence is the fraction of trajectories containing a loop; LBR is computed only over trajectories in which a loop occurs; RA is computed only over trajectories with recovery opportunities.}
\label{tab:outcome_process_comparison}
\begin{tabular}{@{}lrrrrrr@{}}
\toprule
& \multicolumn{3}{c}{Detailed instruction} & \multicolumn{3}{c}{Concise instruction} \\
\cmidrule(lr){2-4}\cmidrule(lr){5-7}
Metric & \multicolumn{1}{c}{Success} & \multicolumn{1}{c}{Failure} & \multicolumn{1}{c}{Gap} & \multicolumn{1}{c}{Success} & \multicolumn{1}{c}{Failure} & \multicolumn{1}{c}{Gap} \\
\midrule
DR & 0.0889 & 0.3067 & +0.2178 & 0.1412 & 0.3618 & +0.2206 \\
Loop incidence & 0.2066 & 0.6525 & +0.4459 & 0.2748 & 0.6520 & +0.3772 \\
LBR & 0.4257 & 0.6390 & +0.2133 & 0.4840 & 0.6142 & +0.1302 \\
RA & 0.7285 & 0.5668 & -0.1618 & 0.6394 & 0.5918 & -0.0476 \\
\bottomrule
\end{tabular}
\end{table}

\subsection{Instruction Granularity}
\begin{figure}[t]
\centering
\includegraphics[width=\linewidth]{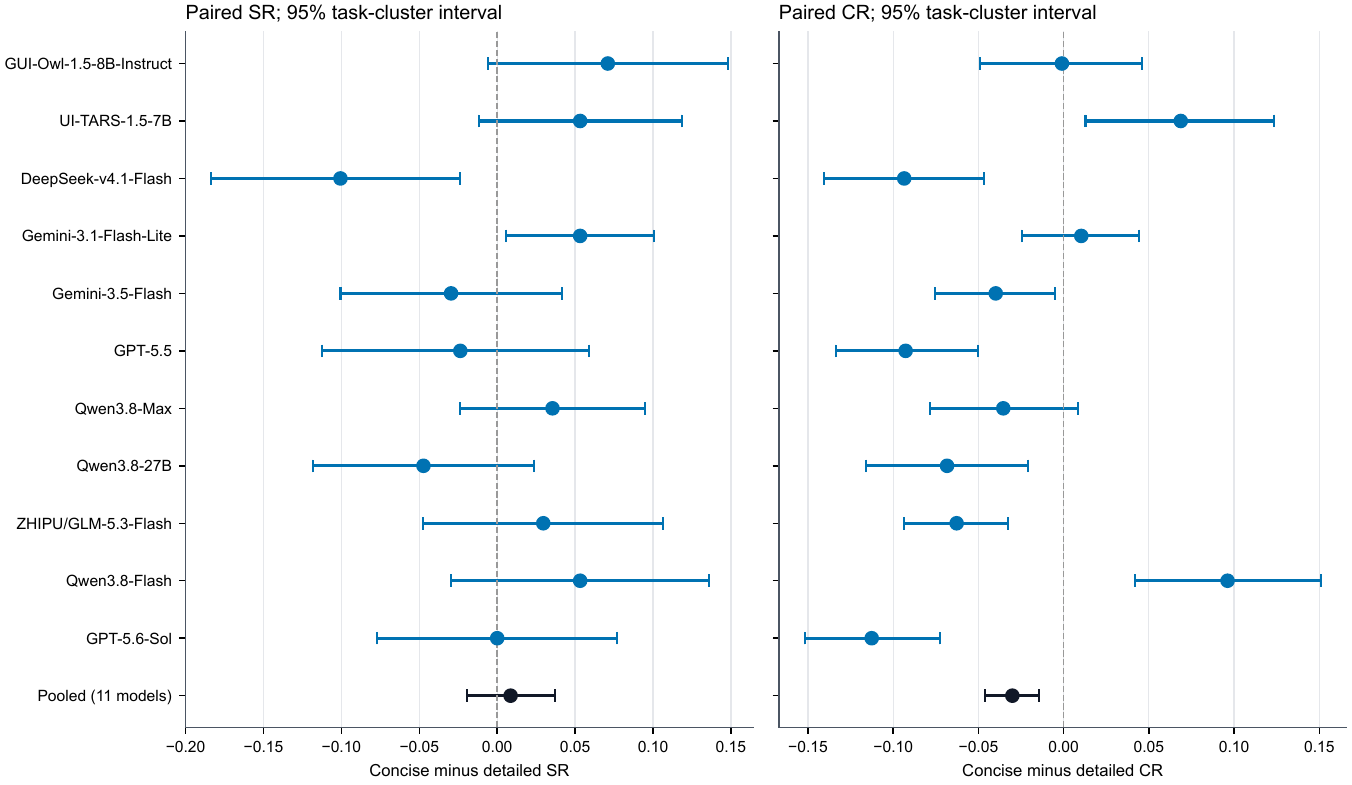}
\caption{\textbf{Instruction effects across all eleven models.} Concise-minus-detailed SR and CR, with task-cluster 95\% intervals. Each model has 169 pairs; the pooled estimate has 1,859 pairs. All 88 replacement runs are selected before pairing.}
\label{fig:complete_instruction}
\end{figure}

\paragraph{Detailed guidance improves path coverage and reduces deviation.}
Across 1,859 matched model--task pairs, detailed and concise instructions yield 639 and 655 successes.
The concise-minus-detailed SR difference is 0.0086 (95\% interval [-0.0194, 0.0371]), whereas the CR difference is -0.0301 (interval [-0.0460, -0.0145]) and the DR difference is 0.0522 (interval [0.0332, 0.0719]).
Thus detailed instructions improve required-path coverage in this cohort, while evidence for an agent-success difference remains inconclusive.
Final-state success and prescribed-path coverage are different estimands: a task can reach its target without completing every reference transition.
Figure~\ref{fig:complete_instruction} shows that effects vary by model, and Table~\ref{tab:instruction_deltas} preserves every reported model's direction.

\begin{table}[t]
\centering\scriptsize
\setlength{\tabcolsep}{4pt}
\caption{Concise-minus-detailed differences from the displayed four-decimal aggregates in Table~\ref{tab:main_results}. Paired intervals in Figure~\ref{fig:complete_instruction} use the complete per-trajectory records for all eleven models.}
\label{tab:instruction_deltas}
\begin{tabular}{@{}lrrrrr@{}}
\toprule
Model & $\Delta$SR & $\Delta$CR & $\Delta$DR & $\Delta$LBR & $\Delta$RA\\
\midrule
UI-TARS-1.5-7B & +0.0532 & +0.0687 & -0.0061 & -0.0561 & -0.0004 \\
GUI-Owl-1.5-8B-Instruct & +0.0710 & -0.0010 & +0.0501 & -0.0518 & -0.0163 \\
DeepSeek-V4.1-Flash & -0.1006 & -0.0936 & +0.0780 & +0.0173 & +0.0418 \\
Qwen3.8-Max & +0.0355 & -0.0354 & +0.0043 & +0.0158 & +0.0850 \\
Qwen3.8-27B & -0.0474 & -0.0684 & +0.0621 & +0.0120 & -0.0293 \\
GLM-5.3-Flash & +0.0296 & -0.0628 & +0.0560 & -0.0181 & -0.0045 \\
Qwen3.8-Flash & +0.0533 & +0.0962 & +0.0551 & -0.1098 & +0.1359 \\
GPT-5.6 Sol & +0.0000 & -0.1127 & +0.0633 & +0.0876 & +0.0071 \\
GPT-5.5 & -0.0237 & -0.0927 & +0.0898 & +0.0652 & -0.0180 \\
Gemini 3.1 Flash Lite & +0.0533 & +0.0103 & +0.0689 & -0.0087 & -0.0673 \\
Gemini 3.5 Flash & -0.0296 & -0.0398 & +0.0532 & -0.0077 & -0.0334 \\
\bottomrule
\end{tabular}
\end{table}

\paragraph{Estimands and pairing.}
All paired differences are concise minus detailed.
CR and DR compare task-level scores; opportunity-dependent LBR and RA are compared only where both members are applicable, and those paired means differ from the pooled denominator-weighted leaderboard estimands.

Among 1,859 pairs, 426 succeed in both conditions, 229 only in concise, 213 only in detailed, and 991 in neither.
The both-success comparisons cover 123 tasks.
Bootstrap samples draw all 169 task IDs with replacement, 10,000 times (seed 20260923), retaining all eligible records for each sampled task.
The intervals use the 2.5th and 97.5th percentiles and condition on this model set; they do not estimate run-to-run device variation.

\begin{table}[t]\centering\small
\caption{Paired process changes when both descriptions succeed. Opportunity-dependent comparisons use their own common support.}
\label{tab:both_success}
\begin{tabular}{@{}lrrr@{}}\toprule
Measure & Pairs & Mean change & 95\% task-cluster interval\\\midrule
DR & 426 & 0.0467 & [0.0236, 0.0713] \\
Loop incidence & 426 & 0.0423 & [-0.0072, 0.0922] \\
LBR & 34 & 0.0339 & [-0.0653, 0.1232] \\
RA & 96 & -0.0911 & [-0.1717, -0.0133] \\
CR & 426 & -0.0591 & [-0.0879, -0.0325] \\
\bottomrule\end{tabular}
\end{table}

\paragraph{Composition and reference-length sensitivity.}
We repeat the pooled paired estimates after excluding each model and each scenario in turn.
The following ranges summarize point estimates, not uncertainty intervals.

\begin{table}[t]\centering\small
\caption{Leave-one-group-out sensitivity of concise-minus-detailed effects.}
\label{tab:composition}
\begin{tabular}{@{}lrrr@{}}\toprule
Excluded unit & SR range & CR range & DR range\\\midrule
One model & [0.0024, 0.0195] & [-0.0427, -0.0218] & [0.0485, 0.0581] \\
One scenario & [-0.0057, 0.0183] & [-0.0365, -0.0153] & [0.0345, 0.0598] \\
\bottomrule\end{tabular}
\end{table}

All 26 composition checks retain a negative CR interval and a positive DR interval; all SR intervals cross zero.
We also split tasks by reference path length at six steps, without choosing the cut to maximize an observed effect.
Shorter paths (at most six steps) have 880 pairs and a CR change of -0.0078 with interval [-0.0329, 0.0167].
Longer paths have 979 pairs and a CR change of -0.0502 with interval [-0.0701, -0.0305].

The longer-path interval favors detailed coverage, whereas the shorter-path interval crosses zero; the difference between subgroup effects is not itself tested.
Scenario estimates are exploratory, have heterogeneous sample sizes, and are not selected as independent confirmatory tests.

\begin{figure}[t]
\centering
\includegraphics[width=\linewidth]{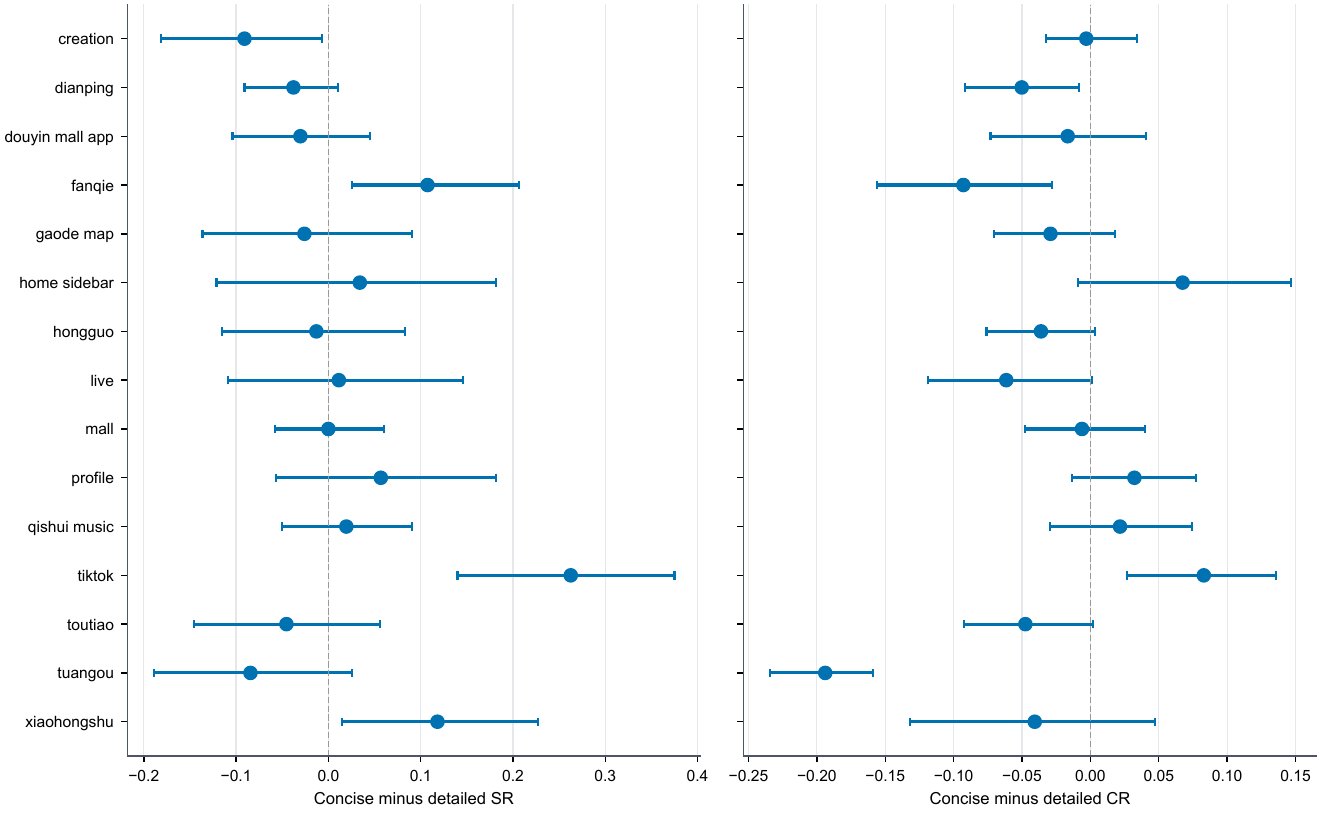}
\caption{\textbf{Scenario-specific paired instruction effects.} All eleven models are included. Intervals resample task clusters; small scenarios have correspondingly less precise estimates. These comparisons are exploratory and unadjusted for multiplicity.}
\label{fig:scenario_effects}
\end{figure}

\subsection{Outcome-Controlled Stability Comparisons}
\paragraph{Outcome-controlled pair construction and sensitivity.}
For each task and condition, let $k$ of the eleven models succeed.
This stratum contributes $\binom{k}{2}+\binom{11-k}{2}$ within-outcome unordered model pairs.
We require a CR gap at most 0.00005 for exact stored-precision matching and require both values of each compared process metric.
We sweep CR tolerances 0.00005, 0.05, and 0.10, and process-gap thresholds 0.10, 0.20, and 0.30.
The resulting 54 outcome--metric--threshold settings use the same source records.
For the primary threshold, task-cluster intervals retain all dependent model pairs from each sampled task.

\begin{figure}[t]
\centering
\includegraphics[width=\linewidth]{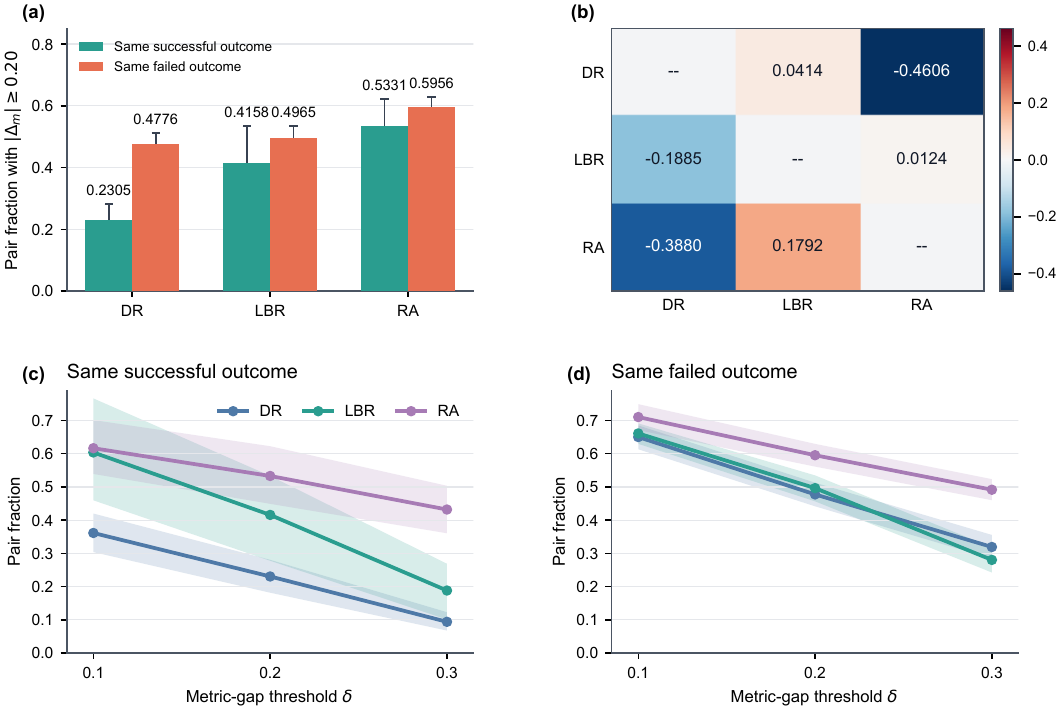}
\caption{\textbf{Outcome-controlled process diagnostics on all 3,718 trajectories.} (a) Exact-CR matched pairs with process gap at least 0.2000; bars distinguish successful and failed pairs and whiskers are task-cluster 95\% intervals. (b) Upper triangle: common-support Spearman correlation; lower triangle: residual correlation controlling task, model, condition, SR, and CR. (c--d) Threshold sensitivity for exact-CR matched pairs; shaded bands show task-cluster 95\% bootstrap intervals.}
\label{fig:diagnostic_matched_pairs}
\end{figure}

\begin{table}[t]\centering\small
\caption{Exact-CR matched pairs with process gap at least 0.2000. Pair counts are not independent sample sizes; intervals cluster by task.}
\label{tab:matched_support}
\begin{tabular}{@{}llrrr@{}}\toprule
Outcome & Metric & Large / eligible & Fraction & 95\% interval\\\midrule
Success & DR & 410/1779 & 0.2305 & [0.1815, 0.2806] \\
Success & LBR & 42/101 & 0.4158 & [0.2778, 0.5333] \\
Success & RA & 153/287 & 0.5331 & [0.4492, 0.6227] \\
Failure & DR & 926/1939 & 0.4776 & [0.4417, 0.5128] \\
Failure & LBR & 425/856 & 0.4965 & [0.4567, 0.5348] \\
Failure & RA & 860/1444 & 0.5956 & [0.5613, 0.6298] \\
\bottomrule\end{tabular}
\end{table}

A stricter descriptive search holds the other two metric gaps at most 0.0500 while requiring the target gap at least 0.2000 and exact CR.
It yields 6 DR pairs, 16 LBR pairs, and 7 RA pairs.
This demonstrates that individual metrics can distinguish some otherwise similar records, without establishing causal isolation.
The common-support correlation analysis includes 1,447 trajectories on 169 tasks; using the same support avoids conflating metric association with different missingness patterns.

\subsection{Cross-task Model Consistency}
\label{app:model_consistency}
Table~\ref{tab:model_consistency} uses all 3,718 records without selecting on SR or CR.
We generate 1,000 partitions with seed 20260926, stratifying by scenario and reference length (at most six versus more than six steps); all models and both instructions for a task remain in the same subset.
Each subset independently aggregates DR over tasks, LBR over loop-containing trajectories, RSR and RA over recovery opportunities, and ARL over successful recoveries.
Unavailable scores are not replaced by zero.
The reported ranges describe sensitivity to task partitioning, not uncertainty from independent repeated experiments.
For the random-ranking baseline, we hold all split-specific rankings fixed and apply one shared random permutation of subset-B model identities across all 1,000 partitions, preserving their dependence.
The one-sided permutation \(p\)-value is the fraction of 200,000 permutations (seed 20260927) whose mean cross-subset Spearman correlation is at least the observed value.
All ten metric--condition tests remain significant after Bonferroni correction.
Restricting RA to episodes initiated by \texttt{off\_path} yields mean rank correlations of 0.4388 and 0.7760 under detailed and concise instructions, respectively, reinforcing the dependence of recovery rankings on opportunity composition.

\begin{table}[H]
\centering
\setlength{\tabcolsep}{3.5pt}
\caption{\textbf{Cross-task reproducibility of model profiles.} Mean Spearman $\rho$ between eleven-model rankings on disjoint task subsets over 1,000 stratified 84/85-task partitions (higher is more reproducible); brackets show the central 95\% of partitions, not confidence intervals. One-sided $p$-values use 200,000 model-identity permutations; all remain significant after Bonferroni correction over ten tests.}
\label{tab:model_consistency}
\begin{tabular}{@{}lcccc@{}}
\toprule
Metric & Detailed \(\rho\) & Permutation \(p\) & Concise \(\rho\) & Permutation \(p\) \\
\midrule
DR & 0.9516 {\scriptsize [0.8636, 0.9909]} & \(<0.0001\) & 0.9228 {\scriptsize [0.8182, 0.9727]} & \(<0.0001\) \\
LBR & 0.9284 {\scriptsize [0.8182, 0.9909]} & \(<0.0001\) & 0.8381 {\scriptsize [0.6727, 0.9636]} & 0.0001 \\
\midrule
RSR & 0.8100 {\scriptsize [0.6182, 0.9364]} & 0.0002 & 0.8269 {\scriptsize [0.7091, 0.9364]} & 0.0002 \\
ARL & 0.5816 {\scriptsize [0.2455, 0.8727]} & 0.0042 & 0.5880 {\scriptsize [0.1998, 0.8636]} & 0.0039 \\
RA & 0.5983 {\scriptsize [0.3000, 0.8818]} & 0.0029 & 0.8508 {\scriptsize [0.6361, 0.9636]} & \(<0.0001\) \\
\bottomrule
\end{tabular}
\end{table}

\subsection{Outcome Relevance of Timely Recovery}
\label{app:recovery_outcome_relevance}
\paragraph{Matched recovery-window analysis.}
We test whether a labeled recovery event is associated with eventual task completion after an actual path deviation.
For each trajectory, we take the first \texttt{off\_path} episode whose zero-indexed onset is at most step 9 and observe the next three decisions, excluding terminal-mismatch starts and isolated recovery labels.
This produces 2,230 eligible trajectories over all 169 tasks.
We match trajectories by task, instruction condition, three-step onset bin, and prior-progress bin, retain strata containing both a return and a non-return, and apply overlap weights.
Among the resulting 1,035 trajectories over 134 tasks, the weighted SR is 0.2591 for a return within three decisions and 0.1275 without one, a difference of 0.1316 with a 95\% task-cluster interval of [0.0800, 0.1829].

\begin{figure}[t]
\centering
\includegraphics[width=\linewidth]{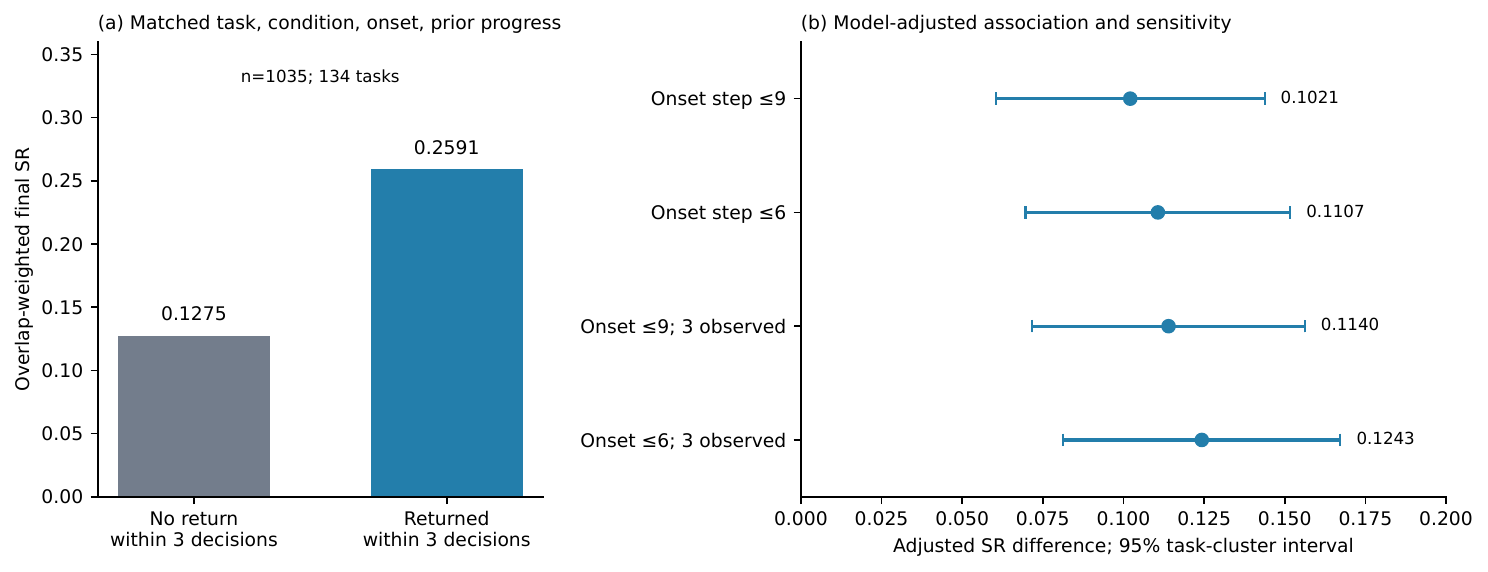}
\caption{\textbf{Timely path return is associated with task success.} (a) Overlap-weighted SR after matching task, instruction condition, deviation-onset bin, and prior-progress bin. (b) Adjusted SR differences under onset and observation-window restrictions; points control task--condition and model fixed effects, onset bin, and prior progress, and whiskers are 95\% task-cluster intervals.}
\label{fig:recovery_outcome_relevance}
\end{figure}

\paragraph{Adjusted association and sensitivity.}
On all 2,230 eligible trajectories, a linear probability analysis controlling task--condition and model fixed effects, onset bin, and prior-progress bin estimates an SR difference of 0.1021 [0.0604, 0.1438].
As Figure~\ref{fig:recovery_outcome_relevance} shows, the estimate remains positive when requiring all three subsequent decisions to be observed or restricting onset to step 6 or earlier.
This consistency connects timely path return to a consequential task outcome, while the observational design leaves naturally occurring error difficulty uncontrolled.

\paragraph{Interpretation boundary.}
Return is neither sufficient for task success nor an isolated validation of the RA latency weighting: among 1,301 trajectories that recovered from every observed \texttt{off\_path} opportunity, only 408 eventually succeeded (SR=0.3136).
After controlling whether a return occurred, a 0.1 increase in the local reciprocal-latency score is associated with an SR difference of 0.0047 [-0.0039, 0.0133].
The experiment therefore supports the outcome relevance of timely return, but not a causal effect, an optimal latency function, or a common-error-start ranking of general recovery competence.

\subsection{Grouped predictive ablation}
\label{app:cost_ablation}
The availability-aware baseline $B$ contains SR, CR, model, instruction condition, scenario, expected task length, graph depth, loop occurrence, and recovery-opportunity occurrence.
The full model is $F=B\cup\{\mathrm{DR},\mathrm{LBR},\mathrm{RA}\}$.
We fit ridge regression with fixed $\alpha=10$ to $\log(1+\mathrm{duration})$ and executed steps.
All 3,718 records have valid target metadata; no success-only selection is applied.
Five-fold cross-validation is repeated over three fixed task-level splits.
All models and both conditions for a task stay in one fold, and categorical encoding and standardization are fitted only on training folds.
An unavailable LBR or RA is encoded as zero only for regression, with availability indicators included on both sides of each comparison.
Intervals resample fixed cross-fitted loss differences by task, without refitting during bootstrap.
They quantify variation over tasks for the fitted procedure, rather than training-instability uncertainty.
Adding all three process metrics reduces cross-validated MSE relative to the availability-aware baseline by 0.1504 for steps and 0.1166 for log duration.

\begin{figure}[t]
\centering
\includegraphics[width=\linewidth]{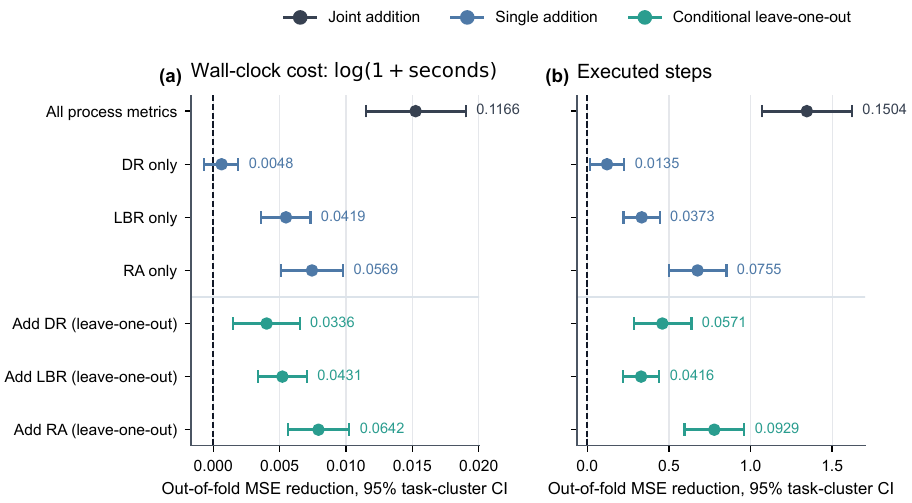}
\caption{\textbf{Grouped predictive ablation on all 3,718 records.} Points show baseline MSE minus augmented-model MSE; positive values favor the augmentation. Whiskers are 95\% task-cluster intervals. Labels give relative MSE reduction as decimals. Blue rows add one metric to the availability-aware baseline; green rows add an omitted metric back to the other two.}
\label{fig:diagnostic_cost_ablation}
\end{figure}

\begin{table}[H]\centering\scriptsize
\caption{Conditional leave-one-out gains. Relative gains are decimals; absolute MSE units differ between targets.}
\label{tab:loo_gains}
\begin{tabular}{@{}llrrr@{}}\toprule
Target & Added metric & Relative gain & MSE reduction & 95\% interval\\\midrule
Log duration & DR & 0.0336 & 0.0040 & [0.0015, 0.0065] \\
Log duration & LBR & 0.0431 & 0.0052 & [0.0033, 0.0071] \\
Log duration & RA & 0.0642 & 0.0079 & [0.0057, 0.0103] \\
Steps & DR & 0.0571 & 0.4599 & [0.2865, 0.6385] \\
Steps & LBR & 0.0416 & 0.3296 & [0.2165, 0.4413] \\
Steps & RA & 0.0929 & 0.7780 & [0.5955, 0.9621] \\
\bottomrule\end{tabular}
\end{table}

All six leave-one-out intervals lie above zero.
The standalone DR duration addition instead has MSE reduction 0.0006 with interval [-0.0007, 0.0019].
This negative result distinguishes standalone utility from conditional complementarity and is retained.

\paragraph{Interpretation and remaining limitation.}
Equal agent success outcomes and path coverage do not determine process quality; matched comparisons, threshold sensitivity, common-support associations, and conditional cost gains provide complementary evidence.
These analyses alone do not validate annotation accuracy, prove statistical independence, or identify uniquely optimal metric definitions; Appendix~\ref{app:empirical_validation} separately audits annotation accuracy on a stratified visual-evidence sample.
Duration can reflect inference speed, device, and collection period; step counts are mathematically related to DR/LBR denominators.
RA is conditional on naturally occurring recovery opportunities and their difficulty, remaining budget, and error mixture.
All bootstrap comparisons are exploratory and unadjusted for multiple testing.
The imported GUI-model package contains evaluated records and timing metadata but excludes media bytes; the present analyses rely on the frozen reported labels.

\begin{figure}[t]
\centering
\includegraphics[width=\linewidth]{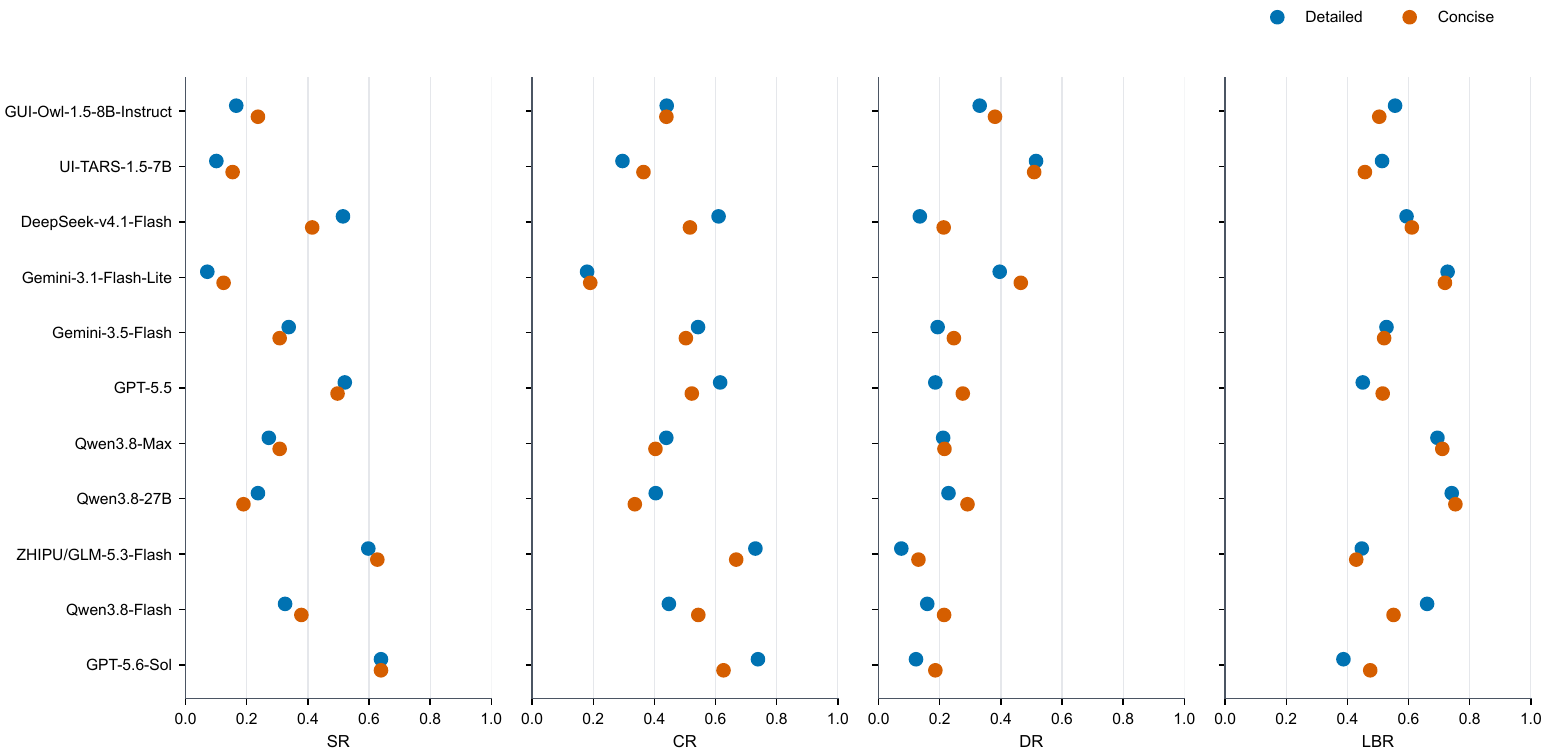}
\caption{\textbf{All eleven models in the full dataset.} Detailed and concise SR, CR, DR, and LBR recomputed from the same 3,718 per-trajectory records used throughout the experiments.}
\label{fig:all_model_overview}
\end{figure}

\section{Automatic UTG Exploration Details}
\label{app:utg_exploration}

\paragraph{Agent--harness decomposition.}
Our explorer is not a fixed-path crawler.
It separates semantic decisions from deterministic device operations: an agent identifies the current GUI state, selects business-relevant actions, and assigns stable state names, while an external harness performs evidence capture, action execution, graph serialization, risk gating, and consistency checks.
This design allows the same execution protocol to cover heterogeneous applications without encoding app-specific navigation paths in the runtime.
Before exploration, we perform a short reconnaissance pass and define an application-specific value spine, such as search--product--cart for shopping, feed--playback--comments for media, or book detail--reader--table of contents for reading applications.
Low-value settings branches are expanded only after the value spine has been covered.

\begin{algorithm}[t]
\caption{Evidence-grounded construction of an application-level UTG}
\label{alg:utg_exploration}
\small
\begin{algorithmic}[1]
\Require Android device \(d\), safety policy \(R\), value-spine targets \(H\)
\Ensure Reviewed UTG \(G=(V,E)\) with state/action evidence
\State Initialize \(G\), pending queue \(Q\), and current state \(v\)
\State \(o \gets \Call{Observe}{d, 2\text{ s}}\)
\While{\(\neg\Call{Stop}{G,Q,H}\)}
    \State \(C_{\mathrm{tap}}, C_{\mathrm{swipe}} \gets \Call{ExtractCandidates}{o.\mathrm{xml}}\)
    \State \(\hat{v} \gets \Call{AgentClassify}{o, C_{\mathrm{tap}}, C_{\mathrm{swipe}}}\)
    \State \(v \gets \Call{MatchOrCreate}{G,\hat{v},o}\); \(\Call{AttachEvidence}{v,o}\)
    \State \(A \gets \Call{AgentPropose}{v,C_{\mathrm{tap}},C_{\mathrm{swipe}},H}\)
    \ForAll{\(a\in A\)}
        \If{\(\Call{ViolatesPolicy}{a,R}\)}
            \State \(\Call{AddBoundary}{G,v,a,\textsc{defer}}\)
        \ElsIf{\(\neg\Call{MountGuard}{v,a}\)}
            \State \(\Call{Block}{a,\textsc{needs-review}}\)
        \Else
            \State \(\Call{AttachPendingEdge}{G,v,a}\); \(\Call{Push}{Q,a}\)
        \EndIf
    \EndFor
    \State \(Q \gets \Call{RankPending}{Q}\)
    \If{\(\Call{NoEligibleAction}{Q}\)}
        \State \textbf{break}
    \EndIf
    \State \(a^\star \gets \Call{AgentSelect}{Q}\); \(\Call{ExecuteOne}{d,a^\star}\)
    \State \(o' \gets \Call{Observe}{d, 2\text{ s}}\)
    \If{\(\Call{Transient}{o'}\)}
        \State \(o' \gets \Call{Observe}{d,\text{extended wait}}\)
    \EndIf
    \State \(v' \gets \Call{AgentVerify}{o',a^\star}\)
    \State \(\Call{ResolveTransition}{G,v,a^\star,v',o'}\); \(\Call{Reconcile}{G}\)
    \State \(o\gets o'\); \(v\gets v'\)
\EndWhile
\State \Return \(\Call{ManualReviewAndDeduplicate}{G}\)
\end{algorithmic}
\end{algorithm}

\paragraph{Observation and evidence capture.}
The default observation waits \(2\) seconds for the interface to stabilize and then records three synchronized evidence sources through ADB~\citep{androiddevelopers2026adb}:
(i) a lossless PNG screenshot using \texttt{exec-out screencap -p};
(ii) the focused package and Activity from \texttt{dumpsys window}; and
(iii) a UI hierarchy from \texttt{uiautomator dump}, with a compressed-dump fallback.
For asynchronous transitions such as opening a reader or a media detail page, the agent requests a second observation with an extended wait rather than treating an intermediate loading frame as a state.
Each observation is stored as a JSON record together with its screenshot and XML paths, timestamp, visible text, candidate controls, scrollable regions, and hashes.
The files are indexed as \texttt{step\_XXXX\_<label>} so every graph update can be traced to the original evidence.

\paragraph{State signatures and matching.}
The XML parser removes invalid regions smaller than \(6\times6\) pixels and deduplicates controls by label and center coordinates quantized to eight-pixel cells.
At most 60 tap candidates, 24 stable controls, and 12 scrollable regions are retained per observation.
Personalized strings containing long numbers, prices, popularity counts, usernames, or other volatile values are excluded from the stable-control set.
The harness computes a 16-hex-character screenshot hash from the PNG and a structural hash from
\[
(\text{Activity},\ \text{node count},\ \text{clickable count},\ \text{first 18 stable controls}).
\]
These SHA-1-based hashes are deterministic identity hints, not security primitives.
The agent first supplies a stable semantic state name; when several nodes share that normalized name, the harness resolves the anchor using exact screenshot hash (\(+160\)), exact structural hash (\(+120\)), matching Activity (\(+100\)), pending status (\(+80\)), and existing evidence/children.
An Activity conflict incurs a \(-40\) penalty.
The final merge decision also checks modal presence, scroll position, selected tab, and semantic role, preventing visually different sheets or scrolled detail states from being merged solely because they share an Activity.

\paragraph{Candidate generation and prioritization.}
For an XML control \(c\), the initial tap score is
\begin{equation}
s_{\mathrm{tap}}(c)=
8\mathbb{I}_{\mathrm{clickable}}+
5\mathbb{I}_{\mathrm{labeled}}+
6\sum_{k\in K_H}\mathbb{I}[k\in \mathrm{label}(c)]
-12\mathbb{I}_{\mathrm{risk}},
\end{equation}
where \(K_H\) contains high-value concepts such as search, playback, comments, filters, lists, reading, and scrolling.
The parser also converts every \texttt{scrollable=true} region into directional swipe candidates.
Vertical regions produce up/down actions, horizontal regions produce left/right actions, and reader pages receive page-turn candidates when XML does not expose a scrollable container.
Swipes are therefore first-class graph edges rather than fallback actions.

The pending-action queue ranks unresolved edges by
\begin{equation}
\begin{split}
s_{\mathrm{plan}}(a)={}&10n_H(a)+w_{\mathrm{priority}}
+18\mathbb{I}_{\mathrm{swipe}}
+4\mathbb{I}_{\mathrm{up/left}}\\
&-100\mathbb{I}_{\mathrm{risk}}
-80\mathbb{I}_{\mathrm{needs\ review}}
-15\mathbb{I}_{\mathrm{recollect}},
\end{split}
\end{equation}
with \(w_{\mathrm{priority}}\in\{30,20,5,-10\}\) for critical, high, normal, and low priorities.
The queue is advisory: the agent selects one action after checking current visual evidence and business value.
The harness then executes exactly one tap, swipe, back, or text-input action.
A swipe lasts 420\,ms and is constrained to the observed scrollable bounds when available; otherwise it uses a screen-relative fallback region.

\paragraph{Graph update and online mount guard.}
In the serialized tree projection, each conceptual UTG edge is represented as
\(\text{state}\rightarrow\text{action record}\rightarrow\text{state}\); in the graph definition, the action remains an edge attribute.
Action records store \texttt{gesture}, \texttt{direction}, \texttt{bounds}, \texttt{priority}, \texttt{status}, and \texttt{risk}; state and action records both retain screenshot, XML, Activity, structural hash, and observation path.
Before a pending edge enters the normal queue, a lightweight mount guard checks the current parent, proposed action, target state, and sibling context.
For example, player-only actions such as comments or episode selection cannot be mounted directly under a search/ranking list, and closing one overlay should not transition directly to an unrelated overlay.
A failed check writes \texttt{mount\_check.status=needs\_review}, sets \texttt{blocked=true}, marks the target \texttt{needs\_reobserve=true}, and removes it from the normal pending queue.
This local guard runs online; a full-tree audit is reserved for completed value-spine milestones and final delivery.

\paragraph{Safety and termination.}
The execution gate uses both explicit metadata and lexical redline categories covering payment, order submission, recharge, withdrawal, authentication, authorization, external sharing, publishing, and account-asset mutations.
A risky entry may be represented as a boundary edge with \texttt{risk=true} and \texttt{status=defer}, but the side-effecting confirmation is not executed.
Search submission and filter application are explicitly treated as safe exceptions.
After each verified transition, reconciliation marks an action \texttt{done} only when every non-risk target has evidence or explored children; blocked, risky, and deferred edges remain unresolved by design.
Exploration stops when the application value spine has been covered and the remaining queue contains only risky, blocked, duplicate, or low-value branches.
Finally, we manually review state names and evidence, deduplicate equivalent states, verify return transitions, and synchronize the canonical graph with its visualization copy.

\subsection{UTG Structural Coverage}
\label{app:utg_structural_coverage}

Table~\ref{tab:utg_structural_coverage} reports an auditable inventory of the final normalized UTGs.
Canonicalization maps 3,265 collected state observations to 2,909 canonical GUI states by merging viewport-only or dynamic-content duplicates while preserving task-relevant differences in modal state, selected tab, scroll position, and semantic role.
An observed transition is a unique \((\text{source state},\text{action label},\text{target state})\) triple for which both endpoints have canonical-state mappings.
We exclude 83 unresolved target observations from the transition counts rather than treating them as verified graph edges.
A branch state has at least two distinct outgoing action--target pairs.

\begin{table}[H]
\centering
\scriptsize
\setlength{\tabcolsep}{3.2pt}
\caption{\textbf{Structural coverage of the ten application-level UTGs.}
States are canonical GUI states; transitions are distinct observed action-conditioned edges; branch states expose at least two distinct outgoing action--target pairs.
Task steps are computed from the 169 curated reference paths and include the cold start as one action.}
\label{tab:utg_structural_coverage}
\begin{tabular}{@{}lrrrrr@{}}
\toprule
Application & Tasks & States & Transitions & Branch states & Mean / max steps \\
\midrule
Douyin & 63 & 827 & 763 & 91 & 6.60 / 11 \\
Dianping & 12 & 206 & 237 & 29 & 9.25 / 10 \\
Douyin Mall & 12 & 268 & 311 & 41 & 6.42 / 12 \\
Fanqie Novel & 11 & 312 & 361 & 38 & 8.00 / 15 \\
Amap & 14 & 171 & 189 & 33 & 7.14 / 9 \\
Hongguo & 14 & 370 & 399 & 47 & 8.93 / 11 \\
Qishui Music & 14 & 145 & 162 & 13 & 6.21 / 9 \\
TikTok & 9 & 163 & 177 & 19 & 7.33 / 13 \\
Toutiao & 10 & 137 & 148 & 24 & 6.20 / 8 \\
Xiaohongshu & 10 & 310 & 369 & 39 & 5.40 / 9 \\
\midrule
\textbf{Total} & \textbf{169} & \textbf{2,909} & \textbf{3,116} & \textbf{374} & \textbf{7.02 / 15} \\
\bottomrule
\end{tabular}
\end{table}

The inventory confirms that every benchmark application is represented by a nontrivial reviewed graph and that the task suite is supported by 3,116 observed transitions and 374 decision-bearing branch states.
Together with the per-task path lengths, these statistics support structural coverage adequacy within GraceBench's declared task and safety scope.
They do not imply exhaustive coverage of all functionality exposed by the underlying applications.
\paragraph{Task-level state support.}
We further measure how much local graph structure supports each modeled task.
For task \(\tau\), let \(S_\tau\) be the union of existing canonical GUI states matched to its curated expert path.
Where multiple route witnesses exist, we retain the nonredundant routes that satisfy the instruction's explicit entry constraints.
Action-only endpoints and task states without a corresponding canonical state do not create synthetic nodes.
We define the undirected one-hop closure over the UTG transitions as
\begin{equation}
N_1(S_\tau)=S_\tau\cup
\{v\in V:\exists u\in S_\tau,\ (u,a,v)\in E\ \lor\ (v,a,u)\in E\}.
\end{equation}
This closure includes the matched task states themselves and every distinct predecessor or successor connected by one modeled action.
It measures the local state context available around a task route, rather than asserting that every neighboring state belongs to a legal completion path.

\begin{table}[H]
\centering
\small
\setlength{\tabcolsep}{4pt}
\caption{\textbf{Canonical states supporting each task subgraph.}
Statistics cover all 169 tasks.
Confidence intervals are nonparametric task-bootstrap intervals with 10,000 resamples and seed 20260925.}
\label{tab:task_subgraph_state_support}
\begin{tabular}{@{}lrrrrr@{}}
\toprule
Measure & Mean & 95\% CI & Median & IQR & Range \\
\midrule
Core task states & 5.91 & [5.62, 6.18] & 6 & [5, 7] & [1, 11] \\
Additional one-hop states & 35.19 & [32.88, 37.66] & 34 & [24, 43] & [6, 102] \\
One-hop closure & 41.09 & [38.62, 43.69] & 40 & [30, 49] & [7, 112] \\
Per-task expansion factor & 7.20$\times$ & [6.82, 7.59] & 7.14$\times$ & [5.33, 8.80] & [2.80, 15.50] \\
\bottomrule
\end{tabular}
\end{table}

The matched core of a task contains 5.91 canonical states on average, while including directly connected one-hop context increases this to 41.09 states and adds 35.19 distinct states.
The mean per-task expansion is \(7.20\times\); under this definition, each task is represented by its matched path states plus a larger set of directly adjacent states.
This statistic quantifies local structural support and does not claim exhaustive application coverage.

\section{Human Expert Audit of Generated Transition Labels}
\label{app:empirical_validation}

\paragraph{Transition sampling and review protocol.}
We froze a visual-validation population of 3,042 trajectories containing
34,767 transitions with locally verified MP4 evidence.
The 676 imported aggregate-only records without media bytes were excluded.
With seed 20260926, we sampled by execution model, instruction condition, and
scenario, retaining at least one trajectory per nonempty stratum.
The resulting 313 trajectories contain 3,487 transitions, or 0.1003 of the
eligible transition population.
Reviewers saw the task, expert path, primitive action, and a five-frame ordered
action-aligned storyboard (pre-action, onset, post-action, response, and
settled frames), while model identity, condition, automatic labels, and
terminal outcomes remained hidden.

Expert agents generated the initial diagnostic labels for exact expert-edge
state matches and Deviation, Recovery, and Loop events.
Human experts then independently reviewed the sampled transition storyboards
under blinded conditions and produced reference binary labels for each
diagnostic field.
After the primary reviews were frozen, a human expert adjudicated field-level
disagreements; all 3,487 transitions were resolved.
Human experts served as auditors and annotators rather than experimental
subjects, so this audit is not a human-subject experiment.

Cohen's $\kappa$ is computed between the expert-agent generated binary labels
and the human-adjudicated reference labels.
For each diagnostic field, the expert-agent output and the human-adjudicated
reference each provide a binary label for every one of the 3,487 sampled
transitions.
The adjudicated labels are used as the human reference to measure agreement
between generated and human expert labels.

\begin{table}[H]
\centering
\small
\setlength{\tabcolsep}{8pt}
\caption{\textbf{Agreement between generated and human expert labels.}
Cohen's $\kappa$ compares expert-agent generated binary labels with
human-adjudicated reference labels over the 3,487 sampled transitions.}
\label{tab:transition_reviewer_agreement}
\begin{tabular}{@{}lc@{}}
\toprule
Field & Cohen's $\kappa$ \\
\midrule
State match & 0.7980 \\
Deviation & 0.7952 \\
Recovery & 0.7663 \\
Loop & 0.8656 \\
\bottomrule
\end{tabular}
\end{table}

\end{document}